\documentclass[]{interact}

\usepackage{epstopdf}
\usepackage{subfigure}
\usepackage{stmaryrd}
\usepackage{enumitem}
\usepackage{algorithm}
\usepackage{algpseudocode}
\usepackage{ bbold }

\usepackage{natbib}
\usepackage{hyperref}
\usepackage{tablefootnote}

\usepackage{lineno}

\bibpunct[, ]{(}{)}{,}{a}{}{,}
\renewcommand\bibfont{\fontsize{10}{12}\selectfont}

\graphicspath{ {./figures/content} }

\theoremstyle{plain}

\theoremstyle{definition}

\theoremstyle{remark}

\begin{document}

\articletype{ARTICLE}

\title{Predicting building types and functions at
transnational scale}

\author{
\name{Jonas Fill\textsuperscript{a,b}\thanks{CONTACT Jonas Fill. Email: jonas.fill@tum.de}, Michael Eichelbeck\textsuperscript{a} and Michael Ebner\textsuperscript{b}}
\affil{\textsuperscript{a}Professur f\"{u}r Cyber Physical Systems, Technischen Universit\"{a}t M\"{u}nchen, Munich, Germany; \textsuperscript{b}Forschungsstelle für Energiewirtschaft e. V., Munich, Germany}
}

\maketitle

\begin{abstract}
Building-specific knowledge such as building type and function information is important for numerous energy applications. However, comprehensive datasets containing this information for individual households are missing in many regions of Europe. For the first time, we investigate whether it is feasible to predict building types and functional classes at a European scale based on only open GIS datasets available across countries. We train a graph neural network (GNN) classifier on a large-scale graph dataset consisting of OpenStreetMap (OSM) buildings across the EU, Norway, Switzerland, and the UK. To efficiently perform training using the large-scale graph, we utilize localized subgraphs. A graph transformer model achieves a high Cohen's kappa coefficient of 0.754 when classifying buildings into 9 classes, and a very high Cohen's kappa coefficient of 0.844 when classifying buildings into the residential and non-residential classes. The experimental results imply three core novel contributions to literature. Firstly, we show that building classification across multiple countries is possible using a multi-source dataset consisting of information about 2D building shape, land use, degree of urbanization, and countries as input, and OSM tags as ground truth. Secondly, our results indicate that GNN models that consider contextual information about building neighborhoods improve predictive performance compared to models that only consider individual buildings and ignore the neighborhood. Thirdly, we show that training with GNNs on localized subgraphs instead of standard GNNs improves performance for the task of building classification.
\end{abstract}

\begin{keywords}
Building types, building functions, graph neural networks, OpenStreetMap
\end{keywords}


\section{Introduction} \label{sec:intro}

With the vision of a carbon-neutral future, energy informatics fields such as power system analysis and energy demand modeling are highly important. Buildings are responsible for 40\% of global energy consumption and 33\% of global greenhouse gas emissions, and ensuring sustainability and energy efficiency for new buildings is key when tackling climate change~\citep{Tricoire2021}. Therefore, the building sector is central to several energy applications. These applications oftentimes require building-specific knowledge such as the number of buildings in an area, building heights, or building types~\citep[p.~13]{Greif}. Categorizing individual buildings according to different functional classes -- for instance, residential and industrial -- and residential building type classes -- for instance, detached and semi-detached houses -- can deliver important information about buildings for applications such as time series of demand in distribution grids~\citep{mueller2017}, energy and heat demand modeling, and for assessing the suitability of buildings for photovoltaics/different kinds of heat pumps~\citep{Greif}.

Accurate building function categorization can also improve synthetic power system test case generation by assigning typical load profiles to each building function type. In contrast, existing approaches typically solely rely on population data for load estimation \citep{zhou2005approximate, birchfield2017grid}. Such test cases can be utilized to develop more robust control approaches and provide the basis for comprehensive benchmarking \citep{eichelbeck2024empowering}.

As software tools such as energy system models tend to consider the entire European continent~\citep{kigle2022}, it becomes increasingly important that building-related information is available for all regions there. Ideally, a class assignment would be readily available for all European buildings. In reality, however, comprehensive datasets do not exist for all countries and regions or policymakers do not release them to the public. In Germany for instance, no nationwide dataset containing the residential typology of individual buildings exists~\citep[p.~36]{Greif}. Even basic datasets that distinguish residential from non-residential buildings are missing in many regions in Europe~\citep{Milojevic-Dupont2023}.

As a consequence of gaps in data, building type/function classification with machine learning techniques has been the subject of many research investigations lately~\citep{Hartmann2024216}. The spatial scopes of the approaches in the literature range from single districts of a city~\citep{Fan2014} to nations~\citep{Sturrock2018, droin2020}. Accordingly, many approaches leverage regionally specific datasets including 3D data~\citep{Henn2012281}, point-of-interest data~\citep{Zhao2023}, or region-specific ground truth~\citep{Xu20222145}, which are not available at a larger scale. To our knowledge, there is no previous study with a European scope.

Many previous studies create estimations by solely considering properties of individual buildings~\citep{droin2020, Huang2017, Sritarapipat201746}. However, buildings do not exist on their own, but are embedded into a socio-economic context~\citep{Fan2014}: certain types of buildings such as detached houses often co-occur within a neighborhood, but industrial buildings are usually far away from residential buildings. One can therefore assume that contextual features such as the form of neighboring buildings, and neighboring structure features such as the distance to neighboring buildings, have substantial predictive power for building types and building functions. The potential for incorporating such features into modeling approaches is likely not yet exhausted. ~\citet{Hartmann2024216} extract the features of neighboring buildings, but not the structure of the neighborhood. Other approaches incorporate aspects of the neighboring structure, but not features of neighboring buildings~\citep{Henn2012281, Atwal2022, Wang2021, Zhao2023}. Few studies utilize the GNN architecture GraphSAGE to perform building categorization and make use of both neighboring features and structure in their input~\citep{Xu20222145, Kong2024, Lei2024}.

However, the previous studies all come with limitations. \citet{Lei2024} only focus on distinguishing residential from non-residential buildings, their approach is therefore not appropriate for energy informatics applications that require more fine-grained classification schemes. Importantly, all former approaches operate on limited spatial scopes, i.e. single cities. According to Tobler's first law of geography~\citep{Fan2014, Tobler1970}, this highly affects the difficulty of the learning task as geometrically similar buildings are more likely to share the same building function if they are locally close. It cannot be readily assumed that statements about model accuracy in these studies are transferable to a diverse set of European buildings. Moreover, all studies operate on a single graph connecting all buildings in the study area to train the GraphSAGE classifier. However, it could be shown that using smaller, localized subgraphs for training is beneficial for model performance~\citep{Zeng2022}. In building type/function classification, there likely is a strong correlation with the properties of directly adjacent buildings, and this correlation gets weaker the further away a building is. We, therefore, hypothesize that classification models should be trained on localized subgraphs around buildings that only contain the closest buildings in terms of distance.

Motivated by these research gaps, it is necessary to develop a new transnational approach for estimating functions and types for buildings where such a semantic attribute is missing. We leverage the OSM database for ground truth data and train our classification models on a randomly selected subset of buildings across the EU, Norway, Switzerland, and the UK. To make use of contextual features, we connect buildings via localized subgraphs and utilize GNNs as classifier models. GNNs can incorporate features of neighboring buildings as well as the structure of the neighborhood.

The main contributions of our study can be summarized as follows:

\begin{enumerate}
    \item We formulate building type and function classification as a transnational task. For this purpose, we construct a multi-source input consisting of datasets available for the entire European continent.
    \item We conduct extensive experiments and show that our multi-source input is suitable for the classification task and that GNNs significantly outperform classical machine learning models that ignore the neighborhood.
    \item We highlight that using localized subgraphs is more suitable for our task than using standard GNNs.
\end{enumerate}

The rest of the article is structured as follows: In section \ref{sec:related_work}, we present related work. Section \ref{sec:datasets} describes the utilized datasets that are used for input and ground truth data. Subsequently, section \ref{sec:graph_neural_network_for_building_classification} defines the semi-supervised learning task, input features, and graph dataset generation. In section \ref{sec:experiments}, we present the experimental results and the conducted analyses. In Section \ref{sec:discussion}, we mention shortcomings and suggest possible future research directions. Section \ref{sec:conclusion} concludes with a summary.

\section{Related work} \label{sec:related_work}

\subsection{Building classification with classical methods}

Building type and building function classification have been tackled in many previous publications. The investigated methods vary in terms of the building classification scheme. Some studies distinguish residential from non-residential buildings~\citep{Sturrock2018, Atwal2022, Zhao2023}, which is useful for applications like population density and distribution estimation. Other authors aim for a more fine-grained classification scheme, including building typology~\citep{droin2020, Henn2012281} or building functions~\citep{Xu20222145, Wang2021}. More fine-grained classification schemes are useful for energy applications such as heat demand prediction~\citep{Greif}. Depending on the type of input features, we identify 4 main groups in literature, which are discussed below:

Remote-sensing-based methods aim to classify buildings by observing aerial images and/or data obtained with LiDAR~\citep{Lu2014134, Huang2017, Sritarapipat201746, Xie20173515}. Remote-sensing-based building classification shows great potential, which is likely not yet exhausted~\citep{Huang2017}. However, it is computationally expensive to work with very high-resolution imagery, which makes the approaches difficult to apply to large regions such as Europe.

Using ground-picture-based methods, it is possible to inspect facade structures next to the roofs of buildings. Building classification based on ground pictures -- oftentimes images from GoogleStreetView -- is studied in several publications~\citep{Kang201844, Zhao2022, Taoufiq20201, Zhang2021}. It has been shown that ground pictures can be an appropriate input source: however, such pictures can be difficult to obtain on a large scale as approaches usually rely on manually downloaded images from GoogleStreetView~\citep{Kang201844,Zhang2021} or on manually collected images~\citep{Taoufiq20201} of a small spatial extract.

Instead of imagery, methods based on GIS (geographic information systems) leverage geographical objects including building footprints~\citep{Henn2012281, Fan2014, Hartmann2024216, Du2015107}, roads~\citep{Sturrock2018, Atwal2022}, point-of-interest data~\citep{Atwal2022}, and land use information~\citep{Atwal2022, Zhao2023} to classify buildings. There exist open GIS datasets available at a global or European scale. OSM~\citep{OpenStreetMap}, a geographic database maintained by volunteers, is available at a global scale and leveraged as a data source in several publications~\citep{Sturrock2018, Atwal2022, Lloyd20201}. Most GIS-based methods rely on some supervised machine learning classifier~\citep{Sturrock2018, Atwal2022, Henn2012281, Xu20222145, Hartmann2024216, Lloyd20201}.

Other approaches for building classification utilize social media images~\citep{Hoffmann2023} or public transportation data~\citep{Zhong2014124}. It has also been shown that it can be effective to leverage a combination of data sources. Examples of such combinations include GIS data, nighttime light, and surface temperatures~\citep{Wang2021}, aerial images combined with ground pictures~\citep{Hoffmann2019}, or aerial images with GIS data~\citep{Du2015107}. A multi-source input consisting of taxi GPS trajectory data, social media data, point-of-interest data, and building footprints from high-resolution images is also leveraged for building categorization~\citep{Niu20171871}.

\subsection{Previous work related to GNNs}

There has been a lot of research interest across scientific fields in recent years related to GNNs. They are applied to the prediction of protein interfaces in chemistry~\citep{Fout20176531}, point cloud classification and segmentation in computer graphics~\citep{Wang2019}, and matrix completion for recommender systems~\citep{Su202151}. In the field of urban form, street networks are analyzed to capture urban form aspects such as the shift from urban to suburban structures~\citep{DeSabbata2023}. GNNs are also utilized to classify groups of buildings~\citep{Yan2019259}.

Few studies tackle predicting building types/functions with GNNs. \citet{Xu20222145} predict building functions for a selection of $\sim$40K buildings in the Chinese city Nanjing, for which they compute a set of 8 geometric shape indicators. \citet{Kong2024} use a multi-source input consisting of geometric, visual, spectral, and socio-economic features to determine the building function of $\sim$10K buildings in the Futian District of the Chinese city Shenzhen. \citet{Lei2024} predict multiple building characteristics in 3 different cities with GNNs. All mentioned studies utilize GraphSAGE as a classifier model and state that this is an appropriate framework for building classification.

\section{Datasets} \label{sec:datasets}

Appropriate datasets are necessary to acquire both input features and ground truth data. The input features are assembled based on multi-source data from OSM and other sources, the ground truth is taken solely from OSM. We require datasets to be available for the entire EU since this makes it straightforward to apply machine-learning techniques. Below we present the included datasets.

\subsection{OpenStreetMap (OSM)} \label{subsec:openstreetmap}

OSM is a volunteered geographic information database maintained by a community of users. It is known as an open map service and contains various geographical objects including buildings, streets, amenities, rivers, land use polygons, administrative boundaries, etc. The database contains 2D footprints of buildings which can be used to extract building geometries as input features. As we assume a form-class relationship, 2D footprint information is highly important.

We also extract country borders from OSM, which we use to encode the countries buildings are located in as an additional input feature. This information is assumed to be predictive since different countries oftentimes have different architectural styles and distributions of building classes.

A subset of buildings further contains a property that describes the typology/function in the \emph{building} tag. Buildings with this semantic attribute are suitable for ground truth data. The tag can hold values to describe the residential typology, e.g. \emph{semidetached\_house} and \emph{terraced}, as well as the building function, e.g. \emph{retail} or \emph{church}. In total, the classification scheme in OSM consists of $\sim$100 building classes, which are mapped to 9 classes relevant for energy contexts: \emph{apartment, detached house, semi-detached house, terraced house}, \emph{industrial building}, \emph{commercial building}, \emph{public building}, \emph{agricultural building}, and \emph{others}. This categorization results in the class imbalance depicted in Fig. \ref{fig:class_distribution_histogram}. The classification scheme is described in more detail in Tab. \ref{tab:building_type_classification_scheme}.

To map building tags to our custom categories, we follow the tag descriptions in the OSM guidelines. Tab. \ref{tab:building_type_mapping_osm} depicts the mapping from OSM tags to our custom classes.
Some building tags in OSM are not mapped to either of our classes and are considered unlabeled. These tags are either not part of the official documentation, not assignable -- for example, it is unclear if a \emph{cabin} is a residential building, not considered buildings, or discarded in favor of more specific tags -- for instance \emph{residential}.

Excluding certain tags further reduces the number of labels in the partially labeled OSM data. After mapping the tags to our classification scheme, the label coverage in the study area is 24.19\%. In Fig. \ref{fig:percentage_osm_labeling}, we note that label coverage varies between different areas, and highlight the importance of filling gaps in the data. We also observe that all European NUTS3 regions have a coverage of at least 1\%, therefore, we assume sufficient diversity in the ground truth data set. We however note that the ground truth data is noisy due to user errors that occasionally occur~\citep{Biljecki2023}.

\subsection{Land use maps} \label{subsec:land_cover_maps}

Land use maps form another part of the input dataset. They consist of polygons that depict different land use classes: a polygon could for example represent a residential area, an industrial area, or a forest. Each building can be assigned to a land-use polygon using a spatial join. We assume that land use maps form an expressive input feature for obvious reasons. Industrial zones typically consist of many industrial and commercial, but few residential buildings. Residential settlements in contrast mostly consist of residential buildings -- but not exclusively.
The CORINE land cover~\citep{EuropeanEnvironmentAgency2019} product is a pan-European land-use map, and the urban atlas~\citep{EuropeanEnvironmentAgency2020} features land-use mapping at higher resolution for 788 European urban areas. To obtain the highest possible accuracy, but still a comprehensive mapping, we combine CORINE land cover and urban atlas: CORINE land cover is used as a backup in areas where no land-use mapping is available in the urban atlas.

Not all thematic classes in the original land-use maps are relevant for building classification. Therefore, we establish a consistent classification scheme by grouping several classes. In total, we identify 15 relevant land use classes. Tab. \ref{tab:lc_classification_scheme} depicts our custom classification scheme, and Tab. \ref{tab:lc_mapping} describes the mapping from the original urban atlas and CORINE land cover classes.

\subsection{Degree of Urbanization} \label{subsec:degree_of_urbanisation}

As an additional part of the multi-source input, we utilize the DEGURBA-classification~\citep{StatisticalOfficeoftheEuropeanUnion2020}. It classifies local administrative units into three types of areas based on population density: \emph{cities}, \emph{towns and suburbs}, and \emph{rural areas}. Using this dataset is motivated by the assumption that certain kinds of areas oftentimes exhibit typical building patterns. For instance, large apartment blocks are most often found in cities, while agricultural buildings are typically located in rural areas. Encoding this information can therefore improve the expressiveness of the input feature set.

\section{Graph Neural Network for Building Classification} \label{sec:graph_neural_network_for_building_classification}

As described in Sec. \ref{sec:related_work}, classical supervised machine learning classifiers are a straightforward way to estimate building classes that were used in many previous studies. To a limited extent, these models can also incorporate information about the neighborhood, for example by taking the average area of all buildings within a certain radius. A more elaborated way of constructing a model input would be a graph dataset. OSM allows to determine spatial relationships between buildings, for example, the distance between two buildings can be computed based on their coordinates. Therefore, we can assemble a graph structure where the centroids of building footprints are the nodes, building attributes such as the area are the node features, and neighboring buildings are connected via an edge -- our definition of neighboring buildings is described in Sec.~\ref{subsec:localized_subgraphs_dataset_generation}. GNNs are deep learning methods that inherently operate on graph input and jointly leverage the graph structure and node feature information. Below we introduce the semi-supervised graph learning task this study is based on.

\subsection{Semi-supervised learning for node-level classification}\label{subsubsec:semi_supervised_learning_for_node_level_classification}

Semi-supervised learning is a machine learning paradigm that can be used for node-level classification in a graph domain. To define it, we introduce the concept of an \emph{undirected graph} $\mathcal{G} = (\mathcal{V}, \mathcal{E})$. It consist of a set of $n$ nodes $\mathcal{V} = \{1,...,n\}$ and a set of $m$ edges $\mathcal{E}$. An edge $e_{ij}$ = $e_{ji}$ is represented by a set $\{i, j\}$ and connects node $i$ and node $j$. Each node $i$ has a \emph{neighborhood} $\mathcal{N}(i) = \{j \in \mathcal{V} \setminus \{i\} \ | \ e_{ij} \in \mathcal{E}\}$. Edge information can be represented in an \emph{adjacency matrix} $\mathbf{A} \in \mathbb{R}^{n \times n}$ defined as

\begin{equation}
   A_{(i, j)} =
    \begin{cases}
        1 & \text{if } e_{ij} \in \mathcal{E}, \\
        0 & \text{otherwise}, \\
    \end{cases} \ \ \text{for } i, j \in \{1,...,n\}.
\end{equation}

In \emph{attributed graphs} $\mathcal{G} = (\mathcal{V}, \mathcal{E}, \mathbf{X}, \mathbf{C}, \mathcal{Y}, \mathbf{Y})$, every node $i \in \mathcal{V}$ has a \emph{node feature vector} $\mathbf{x}_i \in \mathbb{R}^d$ where $d$ is the number of features. We can write these vectors as the following \emph{node feature matrix}:

\begin{equation}
\label{eq:node_feature_matrix}
   \mathbf{X} = \begin{pmatrix} \mathbf{x}_{1}^\intercal \\ \mathbf{x}_{2}^\intercal \\ \vdots \\ \mathbf{x}_{n}^\intercal \end{pmatrix} \in \mathbb{R}^{n \times d}.
\end{equation}

Additionally, every edge $e \in \mathcal{E}$ has an \emph{edge feature vector} $\mathbf{c} \in \mathbb{R}^t$ where $t$ is the number of edge features. $\mathbf{C} \in \mathbb{R}^{m \times t}$ is the \emph{edge feature matrix} containing all edge feature vectors:

\begin{equation}
\label{eq:edge_feature_matrix}
   \mathbf{C} = \begin{pmatrix} \mathbf{c}_{1}^\intercal \\ \mathbf{c}_{2}^\intercal \\ \vdots \\ \mathbf{c}_{m}^\intercal \end{pmatrix} \in \mathbb{R}^{m \times t},
\end{equation}

Further, there is a set $\mathcal{Y} = \{1,...,k\}$ of classes. The data is partially labeled, therefore each node is mapped to either one or no class label: formally, we define such a mapping with a matrix

\begin{equation}
\label{eq:node_feature_matrix}
   \mathbf{Y} = \begin{pmatrix} \mathbf{y}_{1}^\intercal \\ \mathbf{y}_{2}^\intercal \\ \vdots \\ \mathbf{y}_{n}^\intercal \end{pmatrix} \in \{0,1\}^{n \times k},
\end{equation}
where $\forall i \in \{1,...,n\}: \ \| \mathbf{y}_{i}\|_1 \in \{0,1\}$, and $\| \mathbf{y}_{i}\|_1 = 1$ if and only if a label exists for node $i$.

In addition to the attributed graph notation, the semi-supervised setting requires a \emph{training set}, a \emph{validation set}, and a \emph{test set}. In the training phase, a classifier only has access to a subset of class labels. The remaining labels are held out and form the validation and test set, which are used for hyperparameter optimization, and final evaluation respectively. To define training, validation, and test set, we consider subsets of nodes $\mathcal{V}_{\text{train}}, \mathcal{V}_{\text{val}}, \mathcal{V}_{\text{test}}$, and the masks $\mathbf{M}_\text{train}, \mathbf{M}_\text{val}, \mathbf{M}_\text{test}$ defined via

\begin{equation}
\label{eq:masks}
    \begin{split}
        M_{train, (i, j)} & =
        \begin{cases}
            1 \text{ if } i = j, i \in \mathcal{V}_{\text{train}}, \\
            0 \text{ otherwise } \\
        \end{cases} \ \in  \mathbb{R}^{n \times n},\\
        M_{val, (i, j)} & =
        \begin{cases}
            1 \text{ if } i = j, i \in \mathcal{V}_{\text{val}}, \\
            0 \text{ otherwise } \\
        \end{cases} \quad \in  \mathbb{R}^{n \times n},\\
        M_{test, (i, j)} & =
        \begin{cases}
            1 \text{ if } i = j, i \in \mathcal{V}_{\text{test}}, \\
            0 \text{ otherwise } \\
        \end{cases} \quad \in  \mathbb{R}^{n \times n}.
    \end{split}
\end{equation}

The training set, validation set, and test set are defined via matrices $\mathbf{Y}_\text{train}$, $\mathbf{Y}_\text{val}$, and $\mathbf{Y}_\text{test}$ which are given by

\begin{equation}
\label{eq:sets}
    \begin{split}
        \mathbf{Y}_\text{train} & = \mathbf{M}_\text{train} \cdot \mathbf{Y},\\
        \mathbf{Y}_\text{val} & = \mathbf{M}_\text{val} \cdot \mathbf{Y},\\
        \mathbf{Y}_\text{test} & = \mathbf{M}_\text{test} \cdot \mathbf{Y}.
    \end{split}
\end{equation}

The standard supervised learning paradigm is a special case of semi-supervised learning on graphs where $\mathcal{E} = \emptyset$. This corresponds to a graph structure only containing nodes without any edges connecting them. This simplification is useful as it allows us to apply classical machine learning methods such as standard neural networks as baselines for comparison. The individual components of our semi-supervised learning instance are described below.

\subsection{Designing node and edge features for building classification} \label{subsec:designing_node_and_edge_features_for_building_classification}

It is essential to assemble an appropriate set of node features that contains attributes of individual buildings. We construct a set of 69 node features per building and single edge feature, all input features are summarized in Tab. \ref{tab:summary_of_node_and_edge_features}. As 2D building footprints are crucial for predicting building types, we utilize a set of numerical shape indicators that describe the geometry of individual buildings (building-level features). More details about the choice of building-level features are given in Appendix \ref{subsec:building_level_features}. We also incorporate the characteristics of contiguous building blocks such as row houses and connected apartment blocks (block-level features). Further, we include input features describing the land-use class, degree of urbanization, and the building's country.

Based on the assumption that spatially closer buildings influence prediction more, we encode the metric length of edges as a one-dimensional edge feature. The length feature is scaled via an inverse min-max-normalization approach:

\begin{equation}
    \label{eq:minmax_scaling}
    \mathbf{c} = 
    \begin{cases}
        1 - (\textbf{len}_{e} / \xi), & \text{if } \textbf{len}_{e} \leq \xi \\
        0, & \text{if } \textbf{len}_{e} > \xi, \\
    \end{cases}
\end{equation}
where $\textbf{len}_e \in \mathbb{R}$ is the length of edge $e$, $\mathbf{c} \in \mathbb{R}$ is the edge feature vector of edge $e$ -- it is one-dimensional in our case, and $\xi \in \mathbb{R}$ is a threshold value optimized via hyperparameter tuning. This inverse scaling ensures that edges between spatially close buildings have a high edge feature value -- for graph convolutional networks that treat edge features as weights, such a scaling is crucial.

We want to stress that we do not formulate building classification as a label propagation task but as a pure feature propagation task, i.e. we do not encode the partial labels as model inputs. This is because a label propagation approach would likely lead to a model biased toward areas with a high label coverage: it would work well in areas with many labels but might fail to produce sensible predictions in areas with few or no labels. Therefore, it would either require sophisticated evaluation techniques for different label coverage levels or an assumption of sufficient label coverage in the whole study region. Since the former is beyond the scope of the study, and the latter is not fulfilled, we do not include labels as input features.

\subsection{Localized subgraphs dataset generation} \label{subsec:localized_subgraphs_dataset_generation}

Instead of utilizing one large graph for the whole study area, we base training on small, localized subgraphs that are created around buildings. It is possible to create such a subgraph around each building in the study area but to keep training time sufficiently low, we only train our classifier on a selection of $n_\text{graphs}$ subgraphs. The approach for generating the localized subgraphs is twofold:

\begin{enumerate}
    \item We collect nodes for all subgraphs based on the smallest distance to a center node.
    \item We perform Delaunay triangulation~\citep{berg2008} to create edges that connect the nodes.
\end{enumerate}

The generation of the localized subgraphs dataset is discussed in detail below and formalized in Alg. \ref{alg:generate_localized_subgraphs_dataset} in Appendix \ref{sec:dataset_generation}.

\subsubsection{Collecting nodes for localized subgraphs}

To construct localized subgraphs, we follow the approach of randomly sampling $n_\text{graphs}$ labeled nodes from the study area and creating small graphs around these nodes. We define a subgraph $\mathcal{G}_i = (\mathcal{V}_i, \mathcal{E}_i)$ around a node $i \in \mathcal{V}$ by constructing a circular buffer $buf(i)$ with node $i$ as the center. All buildings that intersect this buffer are part of the subgraph:

\begin{equation}
\label{eq:subgraph_construction_criterion}
    \mathcal{V}_i = \{j \in \mathcal{V} \ | \ intersects(building(j), buf(i))\}.
\end{equation}
where $building(j)$ denotes the building geometry belonging to node $j$.

We aim for approximately the same amount of contextual information to be available to each node. However, average distances between buildings vary from area to area -- they are larger in rural areas and smaller in cities. Therefore, the radius of the buffer $buf(i)$ should be a variable function of node $i$ and depend on the number of buildings in the surrounding. To compute the buffer size, we follow an iterative approach: we start with a small circle and enlarge it iteratively until it intersects with some desired number of buildings $n_\text{sub}$ -- we set this variable to 20 as this value proved to be useful in preliminary experiments.

\subsubsection{Creating edges between nodes}

\begin{figure}
\centering
\subfigure[Voronoi cells around centroids of building footprints.]{
\label{fig:delaunay_triangulation_example_a}
\resizebox*{5cm}{!}{\includegraphics{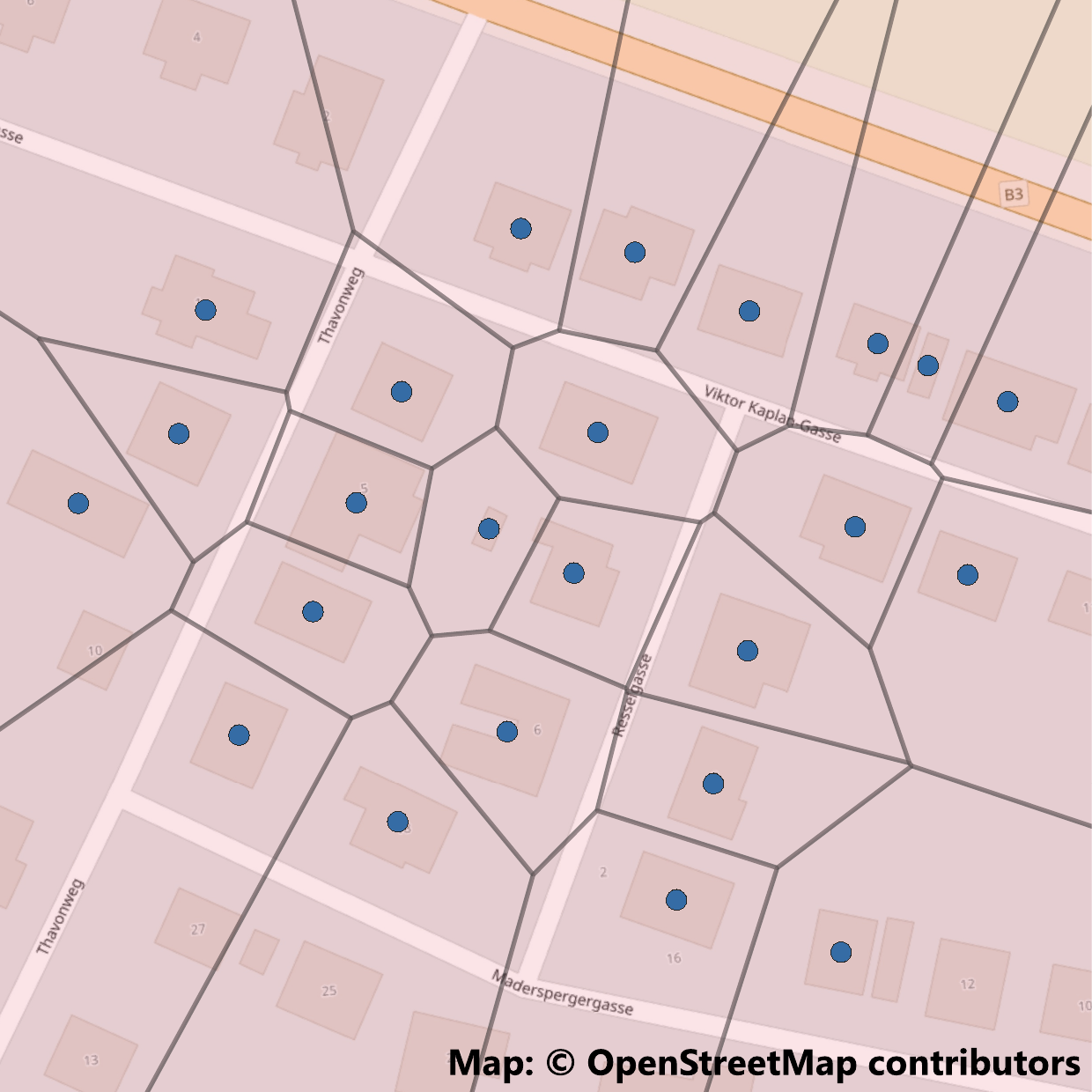}}}\hspace{5pt}
\subfigure[Delaunay triangulation based on the Voronoi cells.]{
\label{fig:delaunay_triangulation_example_b}
\resizebox*{5cm}{!}{\includegraphics{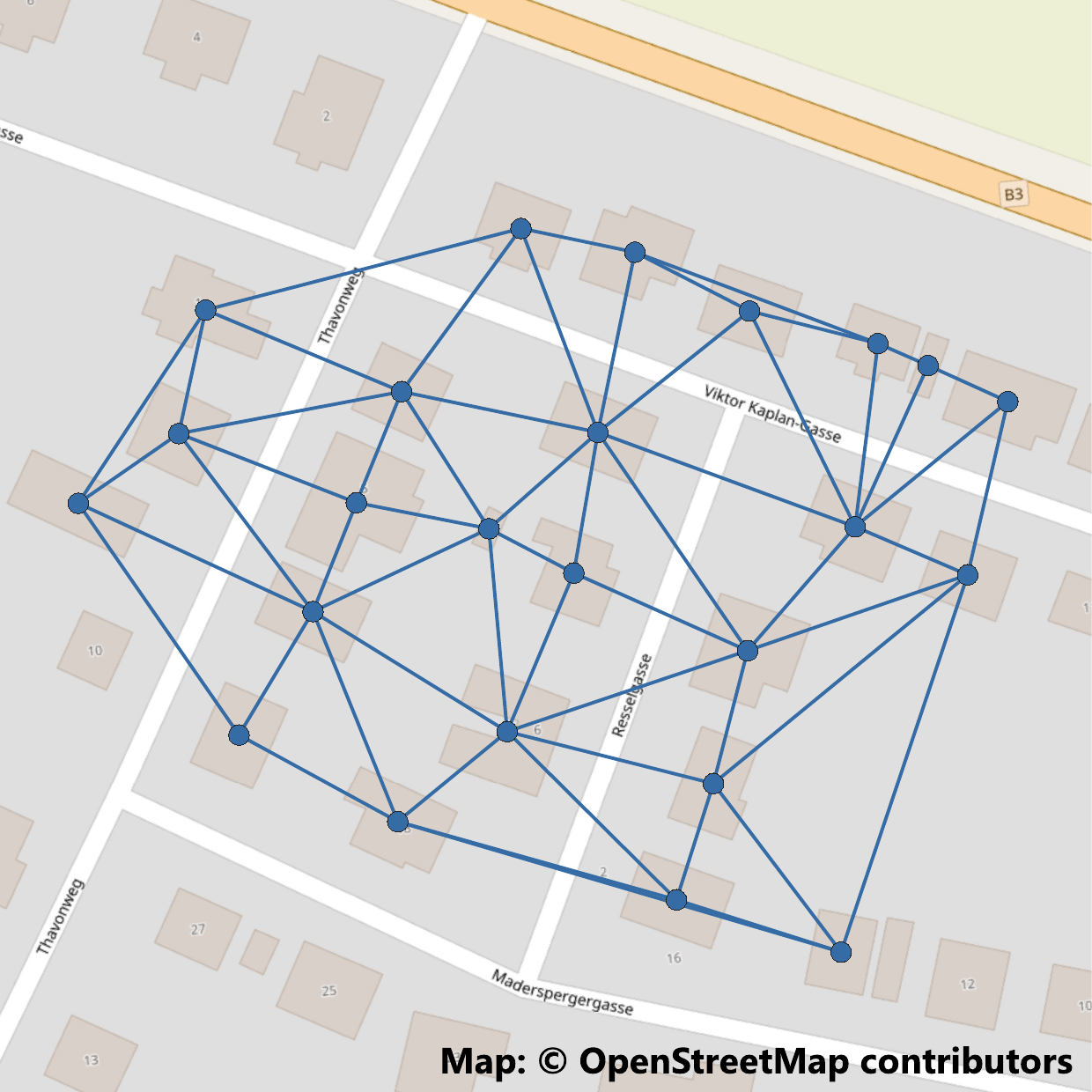}}}
\caption{Triangulation approach to construct graph edges.} \label{fig:delaunay_triangulation_example}
\end{figure}

To create an edge set per subgraph based on the assumption that spatially close buildings influence each other's building classes, we utilize Delaunay triangulation. It is a common method for creating a triangle mesh from a set of points, which is also applied in the building classification context~\citep{Wu20214, Huang2013}. Delaunay triangulation is based on the concept of Voronoi cells in a 2-dimensional Euclidean space~\citep{berg2008}. For an arbitrary node $j \in \mathcal{V}_i$ in subgraph $\mathcal{G}_i$, we denote the corresponding Voronoi cell $\mathcal{R}^{(j)}_i$ as the set of points that are closer to node $j$ than any other node in the subgraph.\footnote{$\mathcal{R}^{(j)}_i = \{x \in \mathbb{R}^2 \ | \ \| x, \mathbf{coord}_j \| \leq \| x, \mathbf{coord}_k \| \text{ for all } k \in \mathcal{V}_i, k \neq j \}$ where $\mathbf{coord}_j \in \mathbb{R}^2$ are the spatial coordinates of node $j$.} By connecting adjacent Voronoi cells with an edge, we obtain the following edge set $\mathcal{E}_i$:

\begin{equation}
    \mathcal{E}_i = \left\{\{j, k\} \ | \ j, k \in \mathcal{V}_i \ \text{and} \ \exists x\in \mathbb{R}^ 2: x\in \mathcal{R}^{(j)}_i \wedge x\in \mathcal{R}^{(k)}_i \right\}.
    \label{eq:neighborhood_condition}
\end{equation}

The approach is visualized in Fig. \ref{fig:delaunay_triangulation_example}.

\section{Experiments} \label{sec:experiments}

In this chapter, we present the experiments conducted in this study. They include a performance comparison between different models, building classes, and subgraph generation methods, an analysis of regional differences, a feature importance analysis, and a comparison with other approaches in the literature.

\subsection{Implementation}\label{subsec:implementation}

\subsubsection{Software tools and hardware} \label{subsubsec:software_tools_and_hardware}

The experimental setup consists of multiple programming frameworks and hardware tools. To store geodata objects such as building polygons, we use \emph{PostgreSQL}~\citep{ThePostgreSQLGlobalDevelopmentGroup.2023} and its geospatial extension \emph{PostGIS}~\citep{PostGISProjectSteeringCommittee.2024}. We store geographical objects in the EU-scale coordinate reference system ETRS89 (EPSG: 3035).

The machine learning experiments are implemented in Python with the deep learning library \emph{PyTorch}~\citep{paszke2017automatic} and its graph-learning extension \emph{PyTorch Geometric}~\citep{FeyLenssen2019} as well as with the machine learning library \emph{Scikit-learn}~\citep{Pedregosa20112825}.

\subsubsection{Performance metrics} \label{subsubsec:performance_metrics}

We use several metrics to evaluate model performances. A straightforward performance metric for classification problems is the overall accuracy. For a test set with $n_{\text{test}}$ samples, ground truth labels $\mathbf{y}_{\text{test}} \in \{0,...,k\}^{n_{\text{test}}}$, and corresponding predictions $\mathbf{\hat{y}}_{\text{test}} \in \{0,...,k\}^{n_{\text{test}}}$, the overall accuracy $OA$ is defined as the share of correct labels:

\begin{equation}
    \label{eq:accuracy}
    OA(\mathbf{\hat{y}}_{\text{test}}, \mathbf{y}_{\text{test}}) = \frac{1}{n_{\text{test}}} \sum_{i = 1}^{n_{\text{test}}} \mathbb{I}({\hat{y}_{\text{test}, i} = y_{\text{test}, i}}) \ \ \in [0, 1],
\end{equation}
where $\mathbb{I}(\cdot)$ is the indicator function. Our dataset has an imbalanced class distribution, see Fig. \ref{fig:class_distribution_histogram}. The accuracy may therefore not reflect the model's performance sensibly. We additionally utilize Cohen's kappa coefficient~\citep{Cohen196037}, a specialized metric for imbalanced datasets. Cohen's kappa coefficient $\kappa(\cdot, \cdot)$ is defined as

\begin{equation}
    \label{eq:cohens_kappa}
    \begin{split}
        p_e(\mathbf{\hat{y}}_{\text{test}}, \mathbf{y}_{\text{test}}) & = \frac{1}{n^2_{\text{test}}} \sum_{k \in \mathcal{Y}} \hat{n}_k n_k, \\ 
        \kappa(\mathbf{\hat{y}}_{\text{test}}, \mathbf{y}_{\text{test}}) & = \frac{OA(\mathbf{\hat{y}}_{\text{test}}, \mathbf{y}_{\text{test}}) - p_e(\mathbf{\hat{y}}_{\text{test}}, \mathbf{y}_{\text{test}})}{1 - p_e(\mathbf{\hat{y}}_{\text{test}}, \mathbf{y}_{\text{test}})} \ \ \in [-1, 1],
    \end{split}
\end{equation}
where $\hat{n}_k$, $n_k$ are the number of times class $k \in \mathcal{Y}$ occurs in prediction vector $\mathbf{\hat{y}}_{\text{test}}$ and ground truth vector $\mathbf{y}_{\text{test}}$ respectively. $\kappa(\mathbf{\hat{y}}_{\text{test}}, \mathbf{y}_{\text{test}}) = 1$ indicated perfect agreement between predictions and ground truth, $\kappa(\mathbf{\hat{y}}_{\text{test}}, \mathbf{y}_{\text{test}}) = 0$ indicates that agreement is not better than chance, and $\kappa(\mathbf{\hat{y}}_{\text{test}}, \mathbf{y}_{\text{test}}) = -1$ indicates disagreement.

\subsubsection{Experimental setup}\label{subsubsec:experimental_setup}

To get a coherent overall picture of the performance different classifier models achieve on the task of building classification, we compare several architectures. The GNN-based classifiers used in this study are the graph convolutional network~\citep{Kipf2017}, GraphSAGE~\citep{Hamilton2017}, the graph attention network~\citep{Veličković2018, Brody2022}, and the graph transformer~\citep{Shi20211548}. As baseline models that only operate on the node features in a supervised manner, we use the decision tree~\citep{Winterfeldt.1986}, the random forest~\citep{Ho1995278}, and the fully connected neural network~\citep{Amari1967299}.

For all classifiers, we consider three different classification tasks, since not all application areas require a fine-granular distinction into 9 building classes:

\begin{enumerate}
    \item \emph{Combined building function and type classification:} all 9 classes
    \item \emph{Residential typology classification:} distinguish between apartment, detached house, semi-detached house, terraced house, non-residential building
    \item \emph{Residential/Non-residential building classification:} distinguish residential from non-residential buildings
\end{enumerate}

It is common in previous literature to only include the center node of localized subgraphs in loss computation~\citep{Zeng2022}. However, computing the loss based on all labeled buildings in the localized subgraphs is a straightforward and efficient way to improve the number of training samples in a large dataset without increasing the number of graphs. We empirically show that this approach leads to improved performance.

Additionally, we examine the effect of localized subgraphs. For this purpose, we train our final model in three settings:
\begin{enumerate}
    \item We construct the localized subgraphs in a distance-based manner as described in Sec. \ref{subsec:localized_subgraphs_dataset_generation}. In this setting, a subgraph always consists of the nodes that are closest to the center node in terms of air-line distance.
    \item We use an alternative way of creating localized subgraphs where subgraphs are crafted out of all nodes that are max. 2 hops in the graph away from a center node. In this setting, our final GNN model with 4 layers can never leverage the complete 4-hop neighborhood.
    \item We do not use localized subgraphs, i.e. with the 4-layer network, the model can always consider the full 4-hop-neighborhood in the graph. There is no limitation in terms of how far the model can 'look' in the graph.
\end{enumerate}

To ensure comparability, we apply the sampling scheme introduced in~\cite{Hamilton2017} to (1). At each hop, a random fixed subset of nodes is sampled to decrease the size of the training batches -- concretely, we use a (3, 3, 2, 2) sampling scheme for our 4 layers. As sampling is only applied to setting (1) and not to (2), we obtain a comparable average number of nodes in training batches. We can alternatively view the setting in (1) as taking large, but sparse subgraphs, and the setting in (2) as taking smaller and denser subgraphs.

The dataset for all experiments consists of 761,274 subgraphs across the study area. For the non-GNN models, we use a flattened version of the graph as described in Sec. \ref{subsubsec:semi_supervised_learning_for_node_level_classification}. For all experiments, we randomly split the subgraphs into graphs for training, validation, and testing. As we aim to fill gaps in the data rather than generalizing to completely unseen regions, such a random split is sufficient, and spatial cross-validation was not necessary. The random split leads to sets of nodes $\mathcal{V}_\text{train}$, $\mathcal{V}_\text{val}$, $\mathcal{V}_\text{test}$ in training, validation, and test set, respectively. As subgraphs can partially overlap, and we use all labels instead of only center labels in general, the setup described in Sec. \ref{subsubsec:semi_supervised_learning_for_node_level_classification} would lead to class labels being in multiple sets. To overcome this, we modify the validation and test masks via

\begin{equation}
\label{eq:masks_modified}
    \begin{split}
        M_{val, ij} & =
        \begin{cases}
            1 \text{ if } i = j , i \in \mathcal{V}_{\text{val}} , i \notin \mathcal{V}_{\text{train}} \\
            0 \text{ else } \\
        \end{cases} \qquad \qquad \quad \ \ \in  \mathbb{R}^{n \times n},\\
        M_{test, ij} & =
        \begin{cases}
            1 \text{ if } i = j , i \in \mathcal{V}_{\text{test}} , i \notin \mathcal{V}_{\text{train}} , i \notin \mathcal{V}_{\text{val}} \\
            0 \text{ else } \\
        \end{cases} \quad \in  \mathbb{R}^{n \times n},
    \end{split}
\end{equation}
and obtain disjoint training, validation, and test sets\footnote{This only applies to labels that do not belong to subgraph center nodes -- center labels are left unchanged.}. To ensure the robustness of the results, we conduct cross validation with 5 different splits into training, validation, and test sets. For each split, we utilize 10 different random seeds. This leads to 50 training runs per classifier, for which average test metrics are reported.

\subsubsection{Parameter selection}\label{subsubsec:parameter_selection}

We perform manual hyperparameter optimization for all classifiers. The results from the hyperparameter optimization step are listed in Tab. \ref{tab:hyperparameters}. We train the final neural network models for a maximum of 200 epochs but apply an early stopping mechanism that terminates training as soon as the validation loss does not decrease for 5 epochs.

\subsection{Results}\label{subsec:results}

\subsubsection{Performance comparison between classifiers}\label{subsubsec:gnns_improve_the_predictive_performance_for_building_classification}

Below we compare the performance between different GNN and non-GNN models for the task of combined building function and type classification. In Tab. \ref{tab:performance_comparison}, we observe that using a graph transformer leads to the highest performance concerning most metrics with a Cohen's kappa coefficient of 0.754 and an overall accuracy of 0.799. Using a graph attention network or GraphSAGE leads to comparable results and significantly higher performance than using non-GNN models. Therefore, we conclude that a GNN-based classifier can successfully leverage contextual features and neighborhood relationships of buildings to improve building class predictions.

\begin{table}
\tbl{Performance comparison between models and classification tasks \\ (combined -- all 9 building classes, typology -- 4 residential typology classes + non-residential class, \newline res./non res. -- only residential + non-residential class). Bold -- highest performance across all classifiers.}
{\begin{tabular}{p{.25\textwidth} p{.2\textwidth} p{.2\textwidth} p{.2\textwidth}}
    \specialrule{.1em}{.05em}{.05em}
    \multicolumn{4}{c}{\textbf{Decision tree}} \\
     & combined & typology & res./non res.  \\
    \hline
    Overall accuracy &0.7236 \tiny \textpm 0.0011&0.8241 \tiny \textpm 0.0010&0.8806 \tiny \textpm 0.0007 \\
    Cohen's kappa &0.6595 \tiny \textpm 0.0013&0.7191 \tiny \textpm 0.0014&0.7581 \tiny \textpm 0.0015 \\
    Range of F1-scores & [0.2413-0.8514] & [0.7366-0.8930] & [0.8648-0.8930] \\
    Macro F1-score &0.6182 \tiny \textpm 0.0015&0.7685 \tiny \textpm 0.0017&0.8789 \tiny \textpm 0.0007\\

    \specialrule{.1em}{.05em}{.05em}
    \multicolumn{4}{c}{\textbf{Random forest}} \\
     & combined & typology & res./non res.  \\
    \hline
    Overall accuracy &0.7484 \tiny \textpm 0.0008&0.8458 \tiny \textpm 0.0008&0.8948 \tiny \textpm 0.0003\\
    Cohen's kappa &0.6898 \tiny \textpm 0.0009&0.7524 \tiny \textpm 0.0008&0.7864  \tiny \textpm 0.0005\\
    Range of F1-scores & [0.2654-0.8617] & [0.7569-0.9064] & [0.8798-0.9064] \\
    Macro F1-score &0.6515 \tiny \textpm 0.0013&0.7994 \tiny \textpm 0.0008&0.8931  \tiny \textpm 0.0002\\

    \specialrule{.1em}{.05em}{.05em}
    \multicolumn{4}{c}{\textbf{Fully connected neural network}} \\
     & combined & typology & res./non res.  \\
    \hline
    Overall accuracy &0.7607 \tiny \textpm 0.0008&0.8556 \tiny \textpm 0.0008&0.9037 \tiny \textpm 0.0004\\
    Cohen's kappa &0.7059 \tiny \textpm 0.0010&0.7671 \tiny \textpm 0.0012&0.8040 \tiny \textpm 0.0008\\
    Range of F1-scores & [0.2951-0.8698] & [0.7697-0.9150] & [0.8889-0.9150] \\
    Macro F1-score &0.6697 \tiny \textpm 0.0018&0.8067 \tiny \textpm 0.0016&0.9019 \tiny \textpm 0.0004\\

    \specialrule{.1em}{.05em}{.05em}
    \multicolumn{4}{c}{\textbf{Graph convolutional network}} \\
     & combined & typology & res./non res.  \\
    \hline
    Overall accuracy &0.7196 \tiny \textpm 0.0017&0.8104 \tiny \textpm 0.0024&0.8544 \tiny \textpm 0.0020\\
    Cohen's kappa &0.6548 \tiny \textpm 0.0022&0.6909 \tiny \textpm 0.0030&0.7023  \tiny \textpm 0.0037\\
    Range of F1-scores & [0.3415-0.8025] & [0.6798-0.8731] & [0.8292-0.8731] \\
    Macro F1-score &0.6493 \tiny \textpm 0.0028&0.7550 \tiny \textpm 0.0025&0.8512  \tiny \textpm 0.0018\\

    \specialrule{.1em}{.05em}{.05em}
    \multicolumn{4}{c}{\textbf{GraphSAGE}} \\
     & combined & typology & res./non res.  \\
    \hline
    Overall accuracy & 0.7902\tiny \textpm 0.0009& 0.8748\tiny \textpm 0.0014& 0.9177\tiny \textpm 0.0010\\
    Cohen's kappa & 0.7438\tiny \textpm 0.001& 0.8000\tiny \textpm 0.0018& 0.8330\tiny \textpm 0.0020\\
    Range of F1-scores & [0.3928-0.8819] & [0.7985-0.9265] & [0.9065-0.9265] \\
    Macro F1-score & 0.7157\tiny \textpm 0.0022&0.8314 \tiny \textpm 0.0017& 0.9165 \tiny \textpm 0.0010\\
    
    \specialrule{.1em}{.05em}{.05em}
    \multicolumn{4}{c}{\textbf{Graph attention network}} \\
     & combined & typology & res./non res.  \\
    \hline
    Overall accuracy &0.7944 \tiny \textpm 0.0010&0.8788 \tiny \textpm 0.0011&0.9203 \tiny \textpm 0.0007\\
    Cohen's kappa &0.7480 \tiny \textpm 0.0013&0.8040 \tiny \textpm 0.0016&0.8375  \tiny \textpm 0.0013\\
    Range of F1-scores & [0.3754-0.8859] & [0.8042-0.9298] & [0.9076-0.9298] \\
    Macro F1-score &0.7181 \tiny \textpm 0.0023&0.8343 \tiny \textpm 0.0020&0.9187  \tiny \textpm 0.0007\\

    \specialrule{.1em}{.05em}{.05em}
    \multicolumn{4}{c}{\textbf{Graph transformer}} \\
     & combined & typology & res./non res.  \\
    \hline
    Overall accuracy & \textbf{0.7989} \tiny \textpm 0.0013 & \textbf{0.8828} \tiny \textpm 0.0013 & \textbf{0.9236} \tiny \textpm 0.0008\\
    Cohen's kappa & \textbf{0.7536} \tiny \textpm 0.0018& \textbf{0.8104} \tiny \textpm 0.0020& \textbf{0.8442} \tiny \textpm 0.0015 \\
    Range of F1-scores & [0.3817 - 0.8884] & \textbf{[0.8087 - 0.9329]} & \textbf{[0.9113 - 0.9329]} \\
    Macro F1-score & \textbf{0.7235} \tiny \textpm 0.0032& \textbf{0.8386} \tiny \textpm 0.0025& \textbf{0.9221} \tiny \textpm 0.0007\\
    \specialrule{.1em}{.05em}{.05em}
\end{tabular}}
\label{tab:performance_comparison}
\end{table}

We, however, observe substantially lower performance for the graph convolutional network. Commonly, graph convolutional networks fail at tasks with low homophily ratios\footnote{The edge homophily ratio of a graph $\mathcal{G} = (\mathcal{V}, \mathcal{E})$ with a label matrix $\mathbf{Y}$ is defined as\\ $h(\mathcal{G}, \mathbf{Y}) = \frac{1}{|\mathcal{E}|} \sum_{\{i, j\} \in \mathcal{E}} \mathbb{1}(\mathbf{y}_i= \mathbf{y}_j),$ where $\mathbb{1}(\cdot)$ is the indicator function.} \citep{Ma2022} and even achieve lower performance than models ignoring the graph structure~\citep{Zhu2020}. In our case, strong homophily is not given in all cases. For instance, garages are not only located next to other garages but are usually surrounded by residential houses. Our results demonstrate that techniques like an attention mechanism -- as used by the graph transformer and graph attention network, and/or a separate weight matrix for the root note -- as it is used in GraphSAGE -- can compensate for the issues related to the graph convolutional network.

\subsubsection{Performance comparison between building classes}

\begin{figure}
\centering
\resizebox*{14cm}{!}{\includegraphics{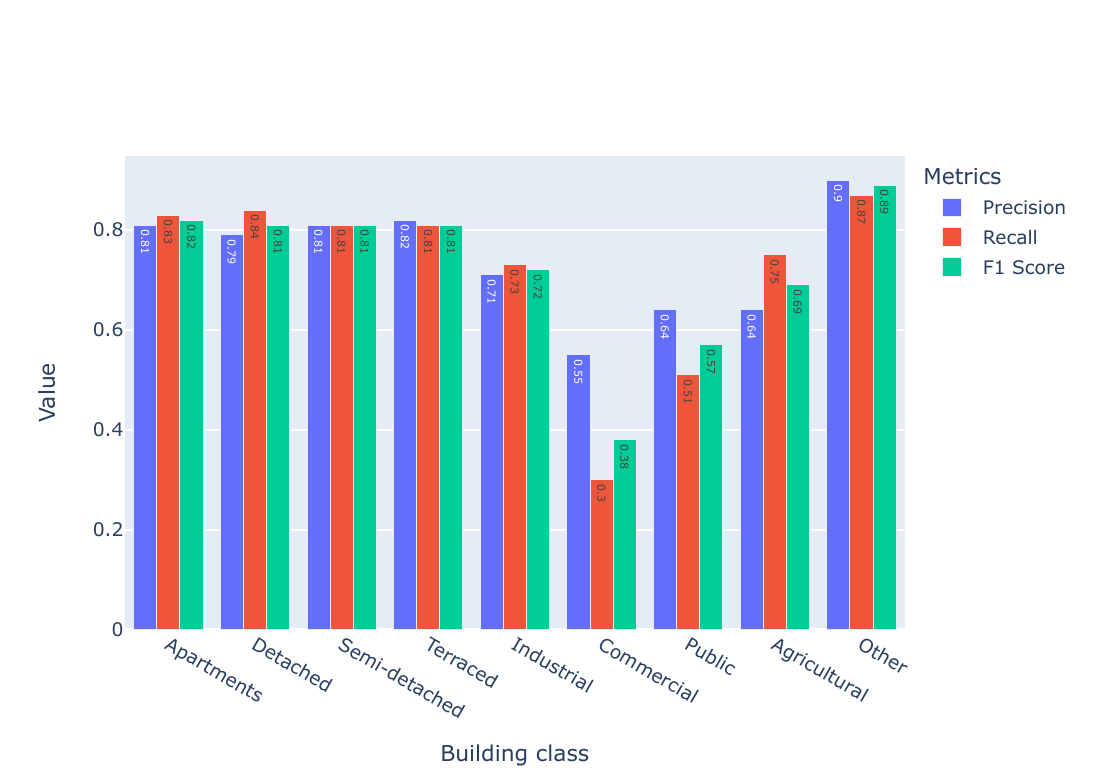}}
\caption{F1 scores, precision, and recall scores of individual building classes for graph transformer.}
\label{fig:f1_score_precision_recall_scores}
\end{figure}

\begin{figure}
\centering
\resizebox*{11cm}{!}{\includegraphics{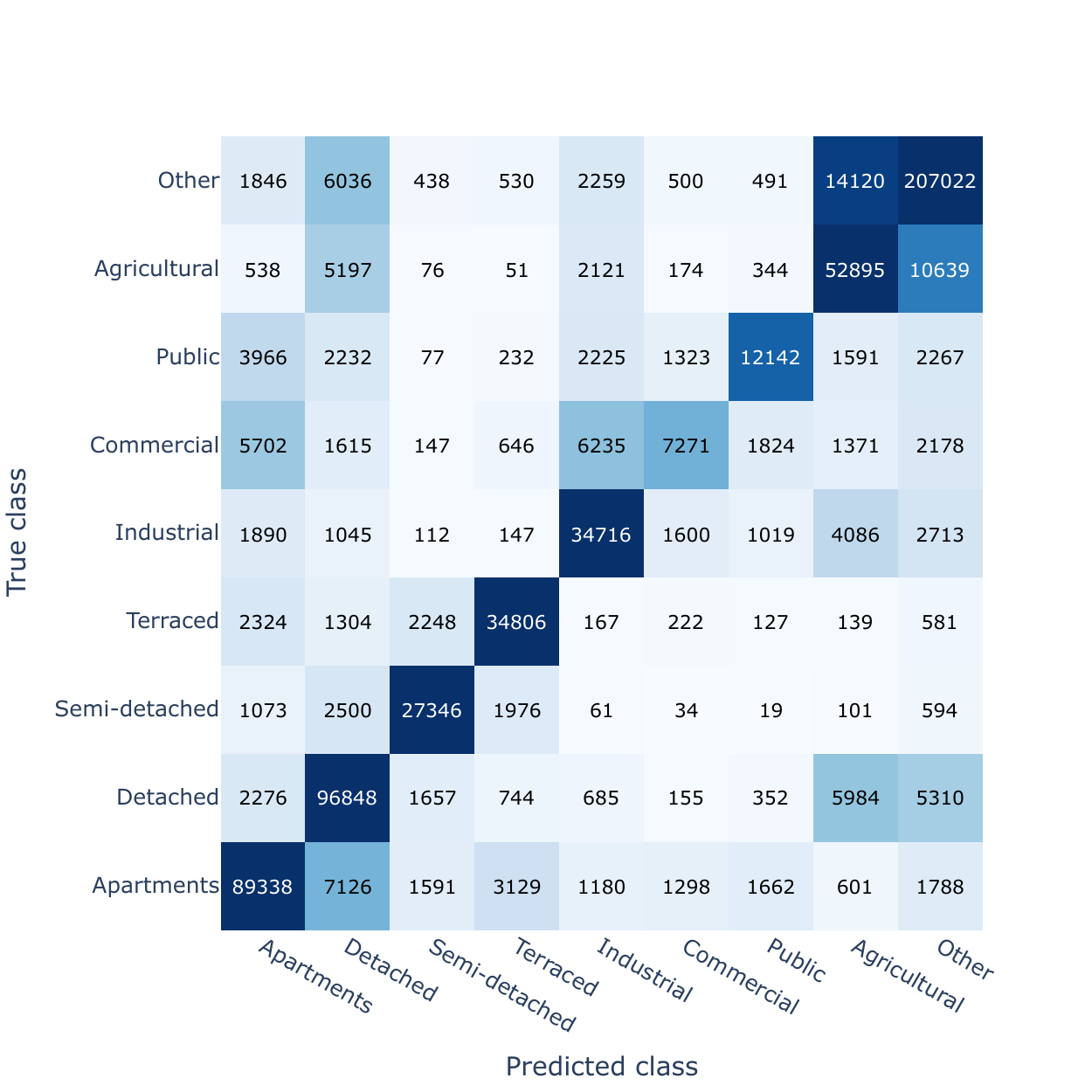}}
\caption{Confusion matrix for graph transformer model.}
\label{fig:confusion_matrix}
\end{figure}

\begin{figure}
\centering
\subfigure[Commercial building \newline predicted as industrial.]{
\label{fig:pred_industrial_gt_commercial}
\resizebox*{4cm}{!}{\includegraphics{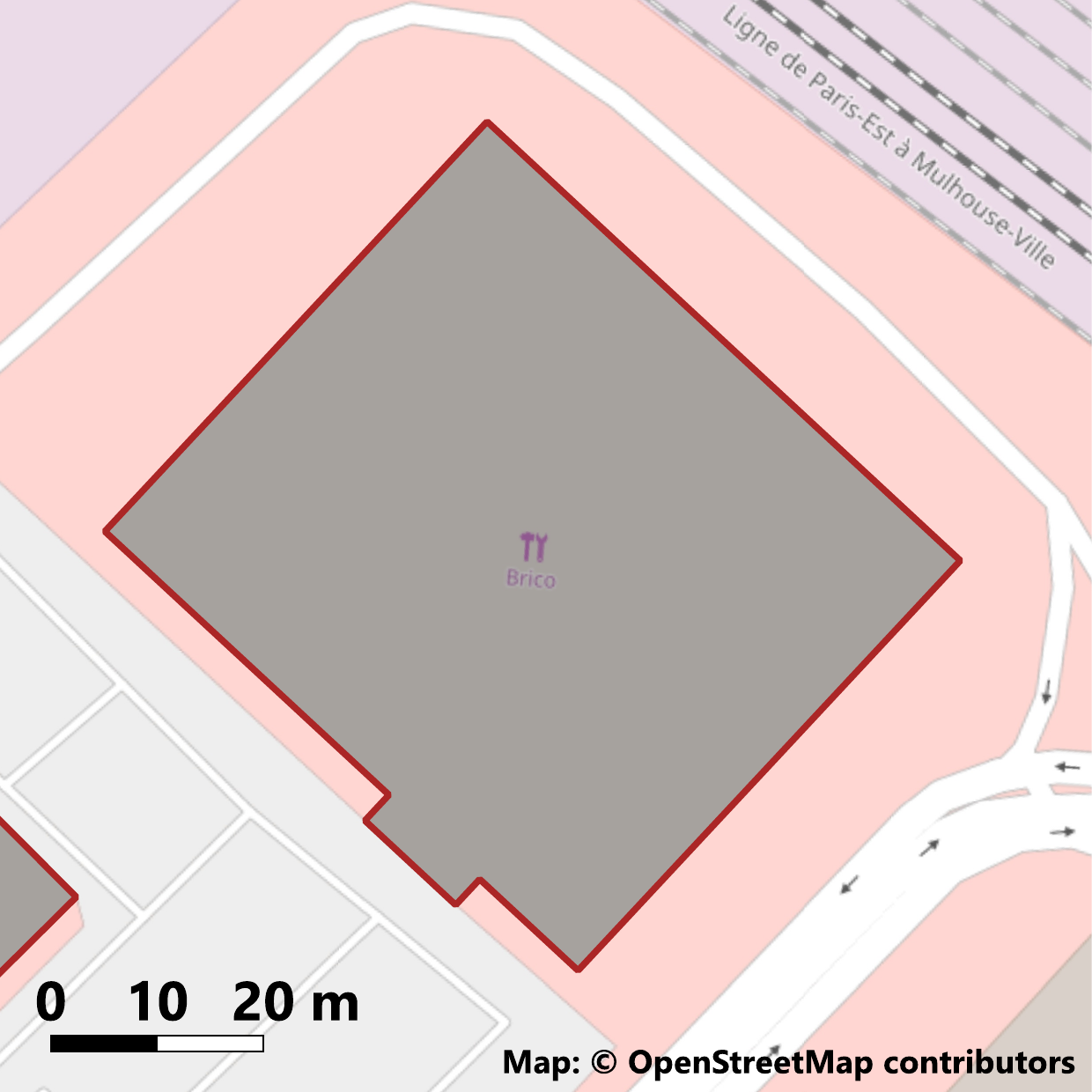}}}\hspace{5pt}
\subfigure[Commercial building \newline predicted as apartments.]{
\label{fig:pred_apartments_gt_commercial}
\resizebox*{4cm}{!}{\includegraphics{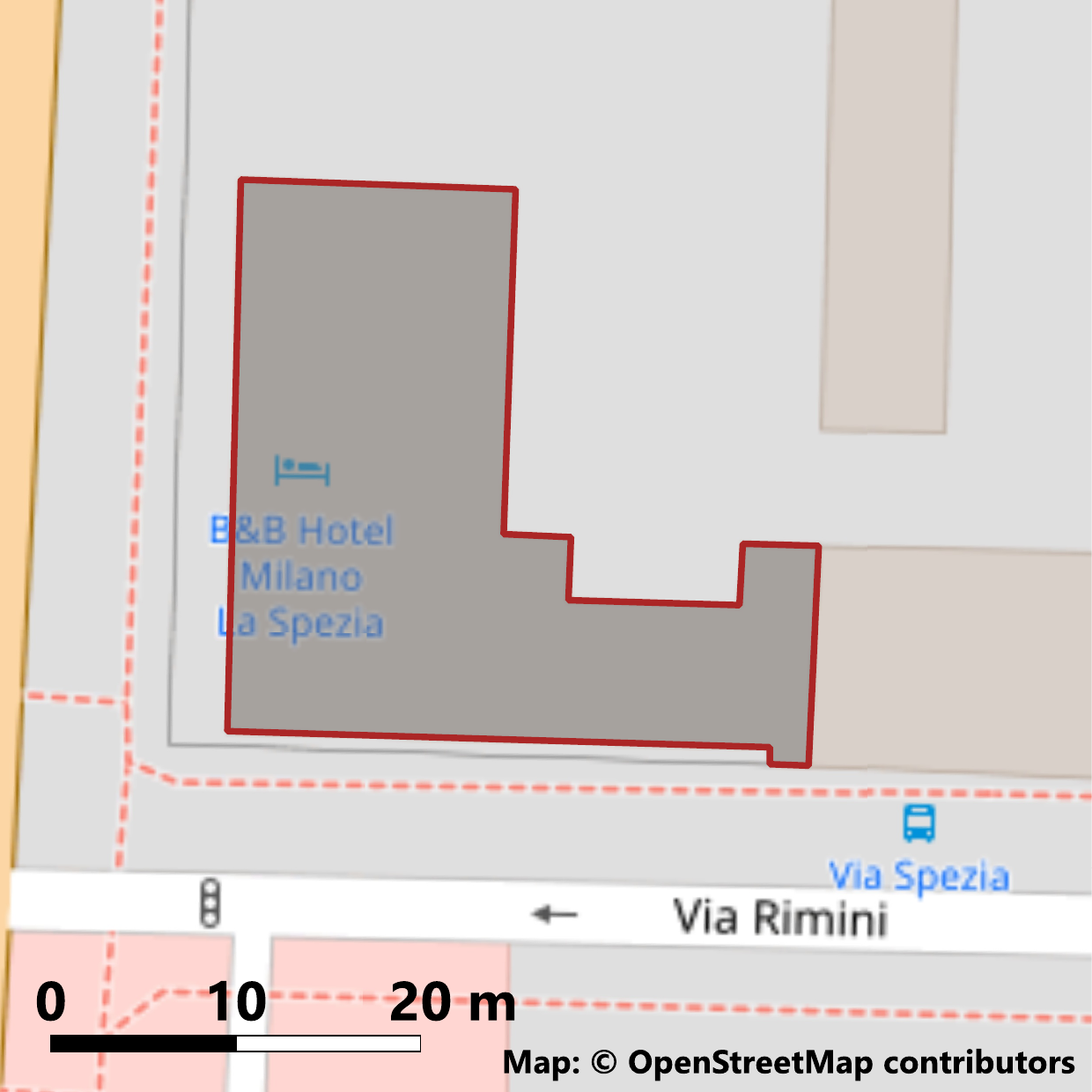}}}\hspace{5pt}
\subfigure[Commercial building \newline predicted as public.]{
\label{fig:pred_public_gt_commercial}
\resizebox*{4cm}{!}{\includegraphics{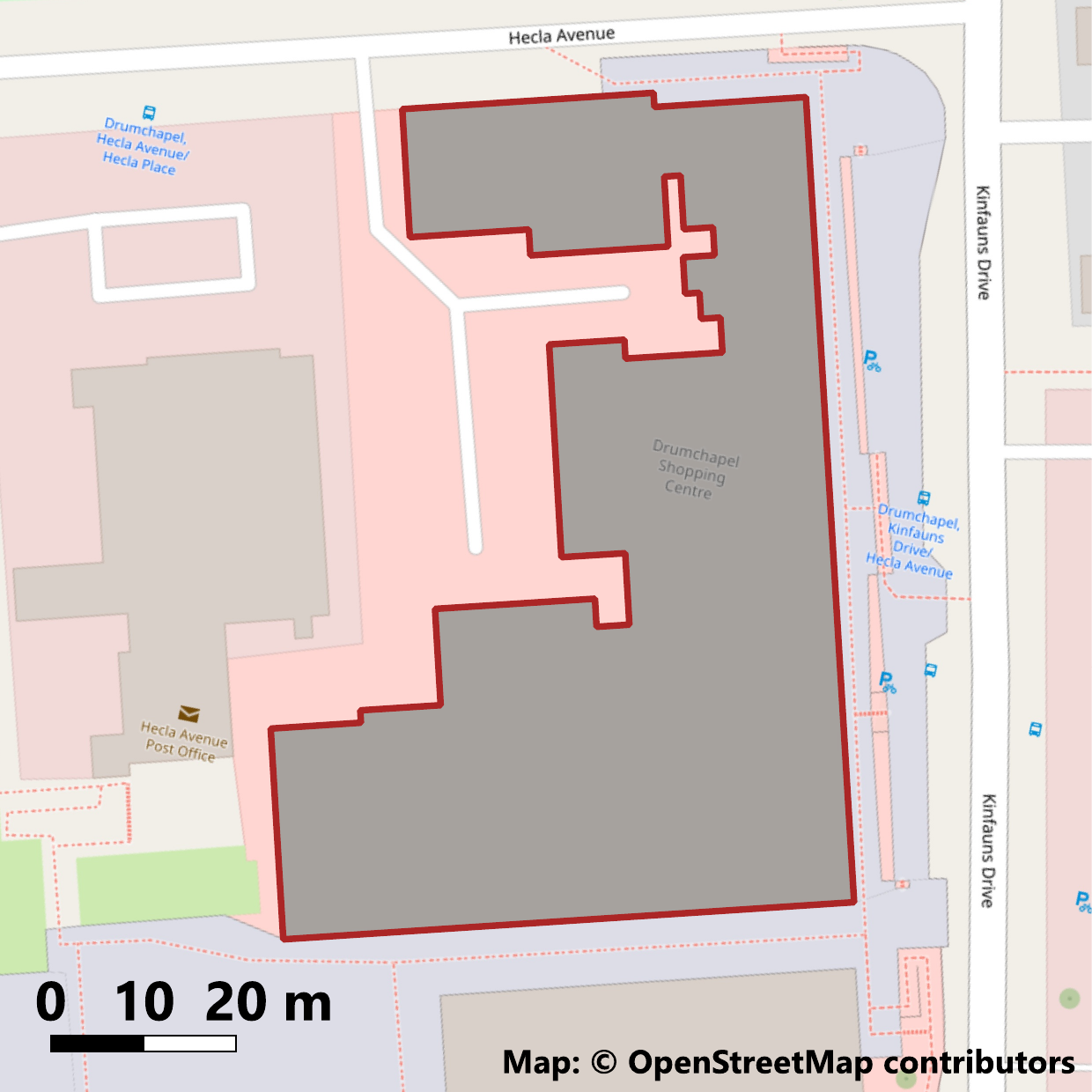}}}\hspace{5pt}
\caption{Examples of typical classification mistakes.} \label{fig:examples_of_typical_classification_mistakes}
\end{figure}

In Fig. \ref{fig:f1_score_precision_recall_scores}, we observe that the graph transformer model gives varying classification performance for the individual classes. For residential building types and the \emph{other}-class, we notice high F1-scores in the range of 0.81 to 0.89. For the non-residential building functions, the F1-scores are lower (0.38 to 0.72). This matches the expectations as residential buildings can often be classified quite clearly based on their shape~\citep{droin2020}, while non-residential buildings such as commercial and public buildings are more difficult to demarcate because of their high diversity in geometric properties~\citep{Fan2014}.

Fig. \ref{fig:confusion_matrix} depicts the confusion matrix based on the $\sim$700,000 labeled buildings in the test set for one representative seed. We perceive a lot of confusion between commercial buildings and industrial buildings. This is not surprising since many commercial buildings have large and simple footprints similar to industrial buildings. Moreover, many commercial buildings are found in industrial zones, therefore, also the land use class is often not decisive. Fig. \ref{fig:pred_industrial_gt_commercial} depicts an example of a large store classified as an industrial building. However, not all commercial buildings are found within demarcated industrial zones, there are also numerous commercial buildings in residential settlements. Certain kinds of commercial buildings including hotels, offices, and smaller stores have similar shapes to apartments -- see example in \ref{fig:pred_apartments_gt_commercial}.

Moreover, a relevant proportion of commercial buildings are misclassified as public buildings. Although public facilities tend to be more complex in shape, we observe that the given input features do not allow a reliable decision in all cases. Fig. \ref{fig:pred_public_gt_commercial} shows an example of a shopping center with a complex shape, similar to typical public facilities such as schools.

\subsubsection{Performance comparison between classification tasks}

For the task of distinguishing residential from non-residential buildings, the graph transformer model achieves very high performance with F1-scores $>$0.9 as shown in Tab. \ref{tab:performance_comparison}. When predicting building typology, we achieve F1-scores $>$0.8, which is still considered high. Further, we notice that the graph transformer, the graph attention network, and GraphSAGE perform significantly better than non-GNN models for the task of residential/non-residential building prediction. The increase of Cohen's kappa coefficient is of a similar order of magnitude as for the combined function and type prediction task. Therefore, we argue that also residential/non-residential building classification is not trivial, but a complex learning task that benefits from more expressive models such as GNNs.

\subsubsection{Performance comparison between subgraph generation methods}\label{ref:performance comparison between subgraph generation methods}

As stated in Sec. \ref{subsubsec:experimental_setup}, we compare two different methods for creating localized subgraphs, and also one variant where a 4-hop neighborhood is leveraged and no localized subgraphs are created. The results are summarized in Tab. \ref{tab:cohens_kappa_comparison_subgraph_generation}\footnote{We use 2 different cross validation splits and 5 random seeds per split in these experiments.}. We observe that constructing localized subgraphs leads to significantly higher Cohen's kappa coefficients w.r.t. all learning tasks than using an unconstrained 4-hop neighborhood. This indicates that directly adjacent buildings carry the most predictive power, buildings further away (e.g. 4 hops) are less relevant. Constructing localized subgraphs with the distance-based approach described in Sec. \ref{subsec:localized_subgraphs_dataset_generation} gives slightly higher performance than using 2-hop subgraphs. This highlights the importance of choosing an appropriate method for creating localized subgraphs. Finally, including all labels in the loss computation -- and not just the center node labels -- further increases performance. This technique can be seen as a way to increase model accuracy at little computational cost rise.

\begin{table}
\tbl{Cohen's kappa coefficient comparison between subgraph generation and label inclusion methods \\ (combined -- all 9 building classes, typology -- 4 residential typology classes + non-residential class, \newline res./non res. -- only residential + non-residential class). Bold -- highest performance across all settings.}
{\begin{tabular}{p{.25\textwidth} p{.15\textwidth} p{.15\textwidth} p{.15\textwidth} p{.15\textwidth}}
    \specialrule{.1em}{.05em}{.05em}
    & 4-hop \newline \tiny(only center label) & 2-hop \newline \tiny(only center label) & distance-based \newline \tiny(only center label) & distance-based \newline \tiny(all labels)  \\
    \hline
    combined &0.6879 \tiny \textpm 0.0027&0.7027 \tiny \textpm 0.0069&0.7131 \tiny \textpm 
    0.0018&\textbf{0.7536} \tiny \textpm 0.0018 \\
    typology &0.7496 \tiny \textpm 0.0032&0.7618 \tiny \textpm 0.0057&0.7701 \tiny \textpm 0.0021&\textbf{0.8104} \tiny \textpm 0.0020 \\
    res./non res. &0.7793 \tiny \textpm 0.0034&0.7907 \tiny \textpm 0.0052&0.7979 \tiny \textpm 0.0021&\textbf{0.8442} \tiny \textpm 0.0015 \\
    
    \specialrule{.1em}{.05em}{.05em}
\end{tabular}}
\label{tab:cohens_kappa_comparison_subgraph_generation}
\end{table}

\subsubsection{Regional performance differences}\label{ref:subsubsec:regional_performance_differences}

\begin{figure}
\centering
\resizebox*{9cm}{!}{\includegraphics{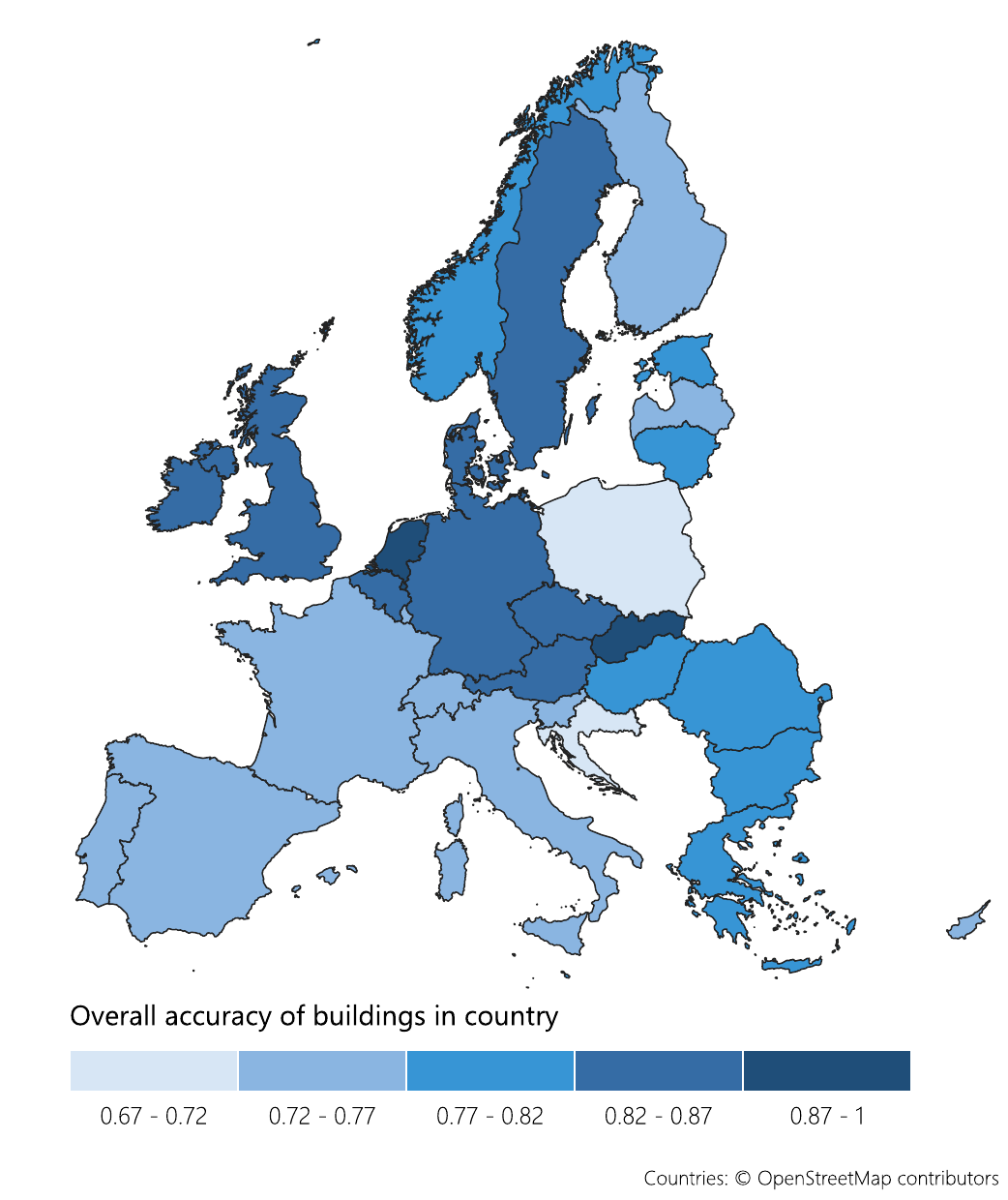}}
\caption{Overall accuracy per country in the study area.}
\label{fig:accuracy_per_country}
\end{figure}

Since the test dataset is randomly distributed across the study area, it is possible to inspect the model performance for individual countries. As depicted in Fig. \ref{fig:accuracy_per_country}, performance is highest in the Netherlands, Belgium, Germany, Austria, Czechia, Slovakia, Denmark, Sweden, the UK and Ireland. Except for Poland and Croatia, we observe an overall accuracy of $>$0.72 in all countries. This shows that the predictive power of the model is not limited to a single region, but that the model can generalize to different countries. Most likely, various effects cause the regional performance differences, for example, missing footprints in OSM, a model biased towards regions with more buildings in the training set, differences in the building distributions, or certain kinds of country-specific buildings that change the difficulty of the learning task.

Besides performance differences between countries, we investigate differences between various degrees of urbanization. We notice that overall accuracy is highest in cities (0.849), followed by towns/suburbs (0.810). Overall accuracy in rural areas is significantly lower (0.732).

\subsubsection{Ablation experiment and feature importance analysis} \label{subsubsec:ablation_experiment_feature_importance_analysis}

We examine the extent our multi-source input improves the predictive performance compared to an input set containing only geometric shape indicators describing individual buildings. For this purpose, we perform an ablation experiment by only using building-level features as input for a graph transformer. The experiment gives a Cohen's kappa coefficient of 0.707, which is substantially lower than the one obtained for the multi-source input.

\begin{figure}[h!]
\centering
\subfigure[Feature importance scores of 10 most \newline influential features.]{
\label{fig:feature_importance_analysis_top_10}
\resizebox*{6cm}{!}{\includegraphics{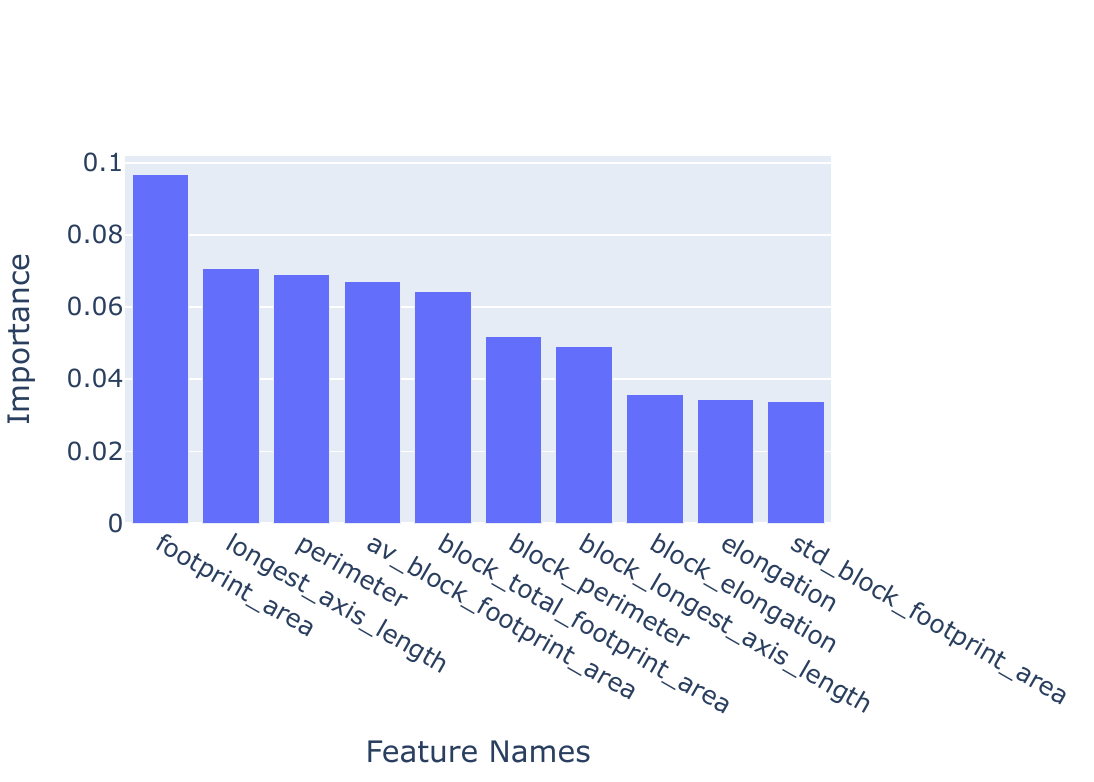}}}\hspace{5pt}
\subfigure[Feature importance scores per group.]{
\label{fig:feature_importance_analysis_groups} \resizebox*{6cm}{!}{\includegraphics{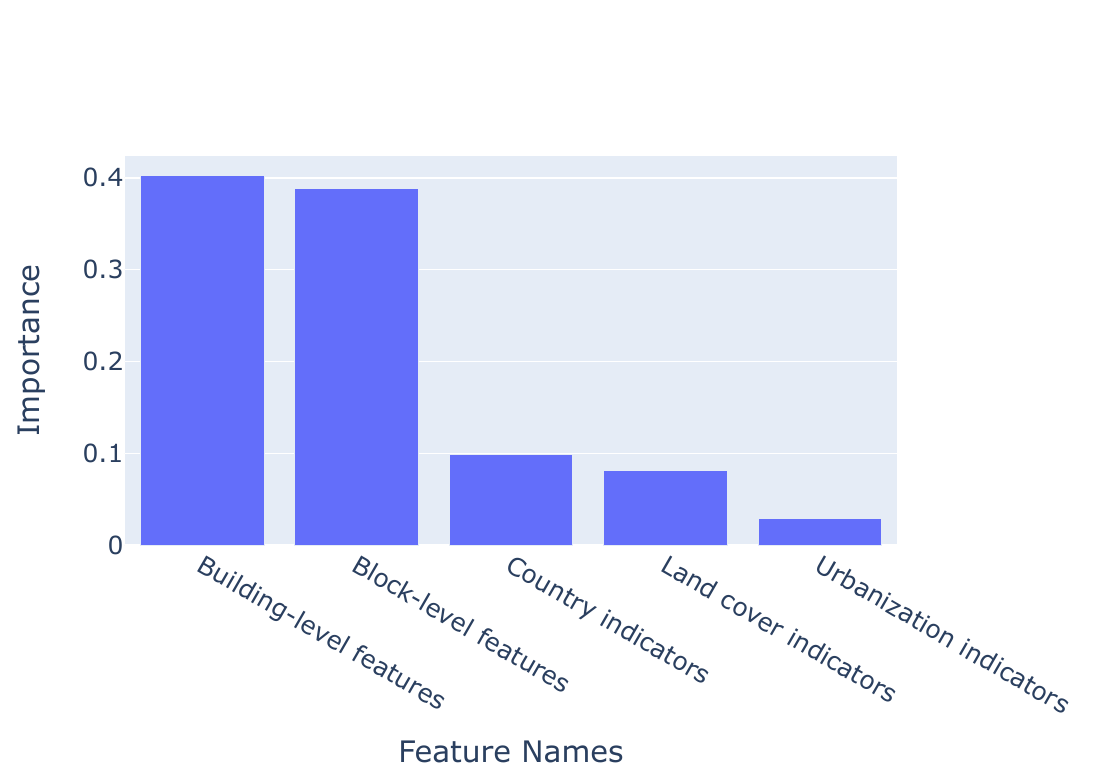}}}
\caption{Feature importance scores based on random forest classifier (averaged over 50 runs with random seeds).} \label{fig:feature_importance_analysis}
\end{figure}

We also perform a feature analysis using the results of the random forest experiment. In Fig. \ref{fig:feature_importance_analysis_top_10}, we observe that area-related features account for  $\sim$26\% of the total importance share. Therefore, we argue that the footprint area alone already has strong predictive power for building classification. But also other geometrical features including the length of the longest axis, perimeter, and elongation carry significant importance.

Fig. \ref{fig:feature_importance_analysis_groups} depicts the distribution of feature importance shares for feature groups. We make out that building-level and block-level features are the most influential, but all feature groups carry some significant predictive power. The high influence of country indicators again highlights that building styles and patterns significantly vary from country to country. The land use class also carries significant importance as it mainly helps to differentiate residential settlements, industrial/commercial zones, and agricultural areas. The degree of urbanization is the least significant when using a random forest classifier.

\subsubsection{Comparison with other approaches for building classification} \label{subsubsec:comparison_with_other_approaches_for_building_classification}

Comparison of our experimental results with the results from other studies is not straightforward due to varying classification schemes, input data, ground truth, and spatial scopes. Nevertheless, we provide a comparison with other approaches to highlight commonalities in the results.

~\citet{Wang2021} perform building function classification on a manually labeled dataset of $\sim$600,000 buildings in Shenzhen, China utilizing geometric shape indicators, point-of-interest data, distances to roads, nighttime light, and land surface temperature measurements as input data. Model-selection-wise, they use a convolutional neural network for embedding the building geometry and gradient boosting for classification. They report rather low F1-scores of 0.38 and 0.52 for commercial and public buildings, respectively, and a high F1-score for residential buildings (0.94), which is comparable to our results.

~\citet{Kong2024} perform building function classification using a multi-source input and GraphSAGE. They also achieve a high F1-score of 0.92 for the residential building class, and lower F1-scores for non-residential building classes like commercial, industrial, and public buildings (between 0.53 and 0.79). Although compatibility with our study is not given in general, we observe the same pattern as in our experiments -- non-residential buildings seem to be more difficult to demarcate.

~\citet{Hartmann2024216} perform classification into residential and non-residential buildings in Germany for a limited, but nationwide dataset. They utilize geometric shape indicators as well as manually computed contextual features of the neighborhood as input features for a random forest classifier. Their approach achieves a Cohen's kappa coefficient of 0.6, which is lower than ours.

~\citet{Zhao2023} classify 380,000 buildings in London into the residential and non-residential classes using OSM for building extraction and ground truth labels. The authors utilize geometric shape indicators, location features such as distances to roads, land use classes, nighttime light, and point-of-interest data as input for a DeepFM model. They achieve an overall accuracy of 0.9, which is comparable to ours.

~\citet{Wu20214} perform building function classification using GraphSAGE. The authors achieve a high overall accuracy of 0.89 and also F1-scores of 0.87 and 0.92 for commercial and public buildings, respectively. These results contrast with other studies that argue that it is difficult to distinguish these classes~\citep{Wang2021, Fan2014}. However, we again highlight that different spatial scopes, input datasets, etc. imply learning tasks of varying difficulty.

\section{Discussion} \label{sec:discussion}

The main objective of this work is to add a building type and function classification approach that can be applied at a transnational scale to the literature. In general, our results show that building type/function prediction for the given scope is feasible, but they also highlight some remaining challenges.

First of all, we point out that the coverage of building footprints in OSM is not complete. Several countries including Bulgaria, Croatia, Hungary, Greece, Spain, Portugal, Romania, and Italy currently show a significant share of missing OSM footprints~\citep{Florio202347}. Creating a comprehensive and pan-European building footprint map based on our approach is therefore not possible using footprints from OSM alone. There exist building footprint conflation approaches on a European scale such as DBSM~\citep{Florio202347} and EUBUCCO~\citep{Milojevic-Dupont2023} that combine footprint data from different data sources. These datasets can be useful to create a more comprehensive map, but also come with limitations -- some data formats in DBSM do not match our expected input data, and EUBUCCO is still not complete for some countries.

To improve on our modeling approach, it could be worth exploring different kinds of graphs with different properties. One such advancement would be a heterogeneous graph that features different kinds of nodes such as roads - an application area for \emph{heterogeneous GNNs}.

Furthermore, we highlight that shape indicators of building geometries were selected manually based on pre-existing work in the field and common knowledge about buildings. More rigorous feature engineering, ablation studies, or an automatic feature extraction approach via convolutional neural networks could give further insides into the most relevant features and potentially improve predictive performance.

The performance differences across the study area indicate that the learning task likely varies from country to country. Although we could show that building classification with European scope is possible, it is worth exploring how the performance of the classifiers changes when only one single country is included in the dataset. As differences in building patterns and architectural styles are usually smaller within a country, we suspect that a model trained on a single country is less biased towards specific regions in the study area.

We also mention that the distribution of building classes in OSM varies from country to country and may not reflect the real distribution of buildings in a region. This effect may lead to a bias of the model towards certain over-represented classes. To overcome such a bias, rigorous regional analyses with additional ground truth sources and regionally-specific models are necessary, which are beyond the scope of this study.

We do not include building height attributes since no open EU-wide dataset containing 3D information on the building level exists. It is however clear that 3D information is predictive for building classes -- they vary in terms of their characteristic heights. The importance of building height information is highlighted in several studies~\citep[e.g.][]{droin2020}. Nevertheless, we achieve sensible performance using our multi-source input and argue that it is not strictly necessary to include 3D information in the input features. For a potential future advancement, we point out that datasets containing 3D information in coarse resolution exist for the entire European continent. It could be worth experimenting with such data as part of the input features.

\section{Conclusion} \label{sec:conclusion}

We predict building types and functions at a transnational scale using ground truth labels from OpenStreetMap and a novel multi-source input dataset assembled from open GIS data. Our semi-supervised graph classifier achieves high performance, especially for the recognition of residential building types. The results are particularly valuable for applications that require building-type information for heterogeneous regions. We show that by using only cross-country datasets such as OSM, it is possible to achieve a Cohen's kappa coefficient of 0.754 for combined typology/function classification and 0.844 for residential/non-residential building classification. These are considered high scores that are comparable to performance metrics obtained in other studies operating with smaller spatial scopes. We show that embedding neighboring features and the graph structure in the model input using GNNs gives significant improvements compared to a fully connected neural network. Further, we demonstrate that creating distance-based localized subgraphs is an effective and performant way to train GNN models for building classification. We argue that graph learning is a promising approach for building categorization that should be investigated in future research.



\section*{Disclosure statement}

The authors declare no conflicts of interest.

\section*{Funding}

This work was partially supported by the German Research Foundation (AL 1185/9-1) and the Bavarian Research Foundation project STROM (Energy - Sector coupling and microgrids).

\section*{Notes on contributors}

\emph{Jonas Fill} received both, his Master’s and his Bachelor's in Informatics from the Technical University of Munich in 2021 and 2024, respectively.
He worked at the geodata lab team at FfE Munich from 2023 to 2024 and wrote his Master’s thesis on predicting building types in Europe in collaboration with FfE and the Professorship of Cyber Physical Systems at the Technical University of Munich.\\ \\
\emph{Michael Eichelbeck} received his Bachelor's in Industrial Engineering from the University of Lüneburg in 2017 and his Master's in Control Systems from Imperial College London in 2019.
Currently, he is a doctoral researcher with the Professorship of Cyber Physical Systems at the Technical University of Munich.
His research focus is on provably safe reinforcement learning, verification of graph neural networks, and contingency-constrained optimal control.\\ \\
\emph{Michael Ebner} received his Master’s in Geography from the University of Heidelberg in 2016. He joined the geodata lab team at FfE Munich in 2018 and has become co-head of it in 2024.
He is currently a doctoral candidate with the Professorship of Renewable and Sustainable Energy Systems at the Technical University of Munich.
His research focus is the application of routing algorithms in the context of energy system research.

\section*{ORCID}

Jonas Fill \url{https://orcid.org/0009-0003-2368-4660} \\
Michael Eichelbeck \url{https://orcid.org/0000-0002-1522-8767} \\
Michael Ebner \url{https://orcid.org/0000-0002-1054-0540}

\section*{Data and codes availability statement}

The data and codes that support the findings of this study are available with the identifier at the private link \url{https://figshare.com/s/fb34b220ed637dcf8b3a}



\bibliographystyle{tfv}
\bibliography{root}

\newpage

\appendix

\section{Input features} \label{sec:input_features}

\subsection{Summary}

The following table summarizes the 70 node and edge input features:

\begin{table}[h]
\tbl{Summary of node and edge features used in this study.}
{\begin{tabular}{ p{.15\textwidth}  p{.13\textwidth}  p{.2\textwidth}  p{.2\textwidth}  p{.09\textwidth} p{.05\textwidth} }
    \specialrule{.1em}{.05em}{.05em}
    \multicolumn{6}{c}{\textbf{Node features}} \\
    Group & Type & Feature names & & Dataset & Count \\
    \hline
    \emph{Building-level} \newline \emph{features} &numerical&
    \begin{itemize}[label=\textbullet, nolistsep, noitemsep, leftmargin=10pt, before*={\mbox{}\vspace{-\baselineskip}}]
        \item footprint area
        \item perimeter
        \item num. corners
        \item anisotropy index
        \item longest axis length
        \item elongation
    \end{itemize}\nointerlineskip&
    \begin{itemize}[label=\textbullet, nolistsep, noitemsep, leftmargin=10pt, before*={\mbox{}\vspace{-\baselineskip}}]
        \item convexity
        \item orientation
        \item num. adjacent \newline buildings
        \item shared wall length
    \end{itemize}\nointerlineskip
    & OSM & 10\\
    \hline
    \emph{Block-level} \newline \emph{features} &numerical&
    \begin{itemize}[label=\textbullet, nolistsep, noitemsep, leftmargin=10pt, before*={\mbox{}\vspace{-\baselineskip}}]
        \item num. footprints
        \item av. footprint area
        \item std. dev. \newline footprint areas
        \item total area
        \item perimeter
    \end{itemize}\nointerlineskip &
    \begin{itemize}[label=\textbullet, nolistsep, noitemsep, leftmargin=10pt, before*={\mbox{}\vspace{-\baselineskip}}]
        \item longest axis length
        \item elongation
        \item convexity
        \item orientation
        \item num. corners
    \end{itemize}\nointerlineskip
    & OSM & 10\\
    \hline
    \emph{Land use} \newline \emph{indicators} &categorical&
    \begin{itemize}[label=\textbullet, nolistsep, noitemsep, leftmargin=10pt, before*={\mbox{}\vspace{-\baselineskip}}]
        \item continuous \newline urban fabric
        \item dense/medium \newline density urban \newline fabric
        \item low density \newline urban fabric,
        \item very low density \newline urban fabric
        \item isolated structure
        \item industrial/ \newline commercial/ \newline public/private
        \item transport
        \item mine dump and \newline construction
    \end{itemize}\nointerlineskip
    &
    \begin{itemize}[label=\textbullet, nolistsep, noitemsep, leftmargin=10pt, before*={\mbox{}\vspace{-\baselineskip}}]
        \item artificial vegetated \newline areas
        \item agricultural
        \item forests
        \item shrub/herbaceous vegetation
        \item open spaces
        \item wetlands
        \item water
        \item urban atlas \newline coverage \newline \emph{(indicates whether \newline area is covered by \newline urban atlas)}
    \end{itemize}\nointerlineskip
    & CORINE land cover, \newline urban atlas & 16\\
    \hline
    \emph{Degree of} \newline \emph{urbanization} \newline \emph{indicators} &categorical&
    \begin{itemize}[label=\textbullet, nolistsep, noitemsep, leftmargin=10pt, before*={\mbox{}\vspace{-\baselineskip}}]
        \item cities
        \item towns/suburbs
    \end{itemize}\nointerlineskip
    &
    \begin{itemize}[label=\textbullet, nolistsep, noitemsep, leftmargin=10pt, before*={\mbox{}\vspace{-\baselineskip}}]
        \item rural areas
    \end{itemize}\nointerlineskip
    & DEGURBA & 3\\
    \hline
    \emph{Country} \newline \emph{indicators} &categorical&
    \multicolumn{2}{p{.4\textwidth}}{one category per country in study area: \newline at, bg, de, etc.}
    & OSM & 30\\
    \specialrule{.1em}{.05em}{.05em}
    \multicolumn{6}{c}{\textbf{Edge features}} \\
    Group & Type & Feature names & & Dataset & Count \\
    \hline
    \emph{Edge length} &numerical&Edge length& & OSM & 1\\
    \specialrule{.1em}{.05em}{.05em}
    \multicolumn{5}{c}{\textbf{Total number of features}} &\textbf{70}\\
    \specialrule{.1em}{.05em}{.05em}
\end{tabular}}
\label{tab:summary_of_node_and_edge_features}
\end{table}

\subsection{Building-level features}\label{subsec:building_level_features}

Building-level shape indicators carry high importance for predicting building types. There are various sensible choices for such features. For example,~\citet{Atwal2022} only use footprint areas to describe the geometry of buildings. We aim for a more expressive set of shape indicators similar to the one introduced by~\citet{Milojevic-Dupont2020}. They utilize a set of 10 shape indicators for building footprints and apply their feature set to infer multiple semantic attributes of buildings, showing great suitability of the feature set~\citep{Milojevic-Dupont2020, Nachtigall2023}. Therefore, we utilize this set of shape indicators for our study. The features are listed below:

\subsubsection{Footprint area, perimeter, and number of corners}

For each building, we compute the footprint area in $m^2$, the footprint perimeter in $m$, and the number of corners as basic geometric features. We define a corner as an angle $\alpha$ that is spanned by 3 points $p_1,p_2,p_3$ subsequently lying on the exterior ring of the geometry. The angle must fulfill

\begin{equation}
    \alpha \leq 170\text{°}, \alpha \geq 190\text{°}
    \label{eq:corner}
\end{equation}
to be considered a corner.

\subsubsection{Anisotropy index, longest axis length, elongation, and convexity}

In order to capture more nuanced geometric and morphological features of building footprints, we use the \emph{anisotropy index} $\phi$, the \emph{longest axis length} $\lambda$, the \emph{elongation} $\eta$, and the \emph{convexity} $\gamma$. We assume that such features hold additional predictive power, which is shown in several studies~\citep{Zhao2023, Xu20222145, Milojevic-Dupont2020}. Table \ref{tab:morphological_features} describes these four shape indicators in more detail.

\begin{table}
\tbl{Anisotropy index, longest axis length, elongation, convexity.}
{\begin{tabular}{p{.16\textwidth} p{.37\textwidth} p{.37\textwidth}}
    \specialrule{.1em}{.05em}{.05em}
    \emph{Feature} & \textbf{Anisotropy index} & \textbf{Longest axis length} \\
    \hline
        \emph{Formula} & 
        \begin{equation*}\phi = \frac{\text{Area}(F)}{\text{Area}(C)}\end{equation*} \newline
        $F :$ footprint geometry \newline 
        $C :$ minimal circumscribed circle around footprint &
        $\lambda =$ diameter of the minimal circumscribed circle around footprint (in m)  \\
    \hline
    \emph{Scale} & [0, 1] & $\mathbb{R}$  \\
    \hline
    \emph{Visualization} & \includegraphics[scale=0.25]{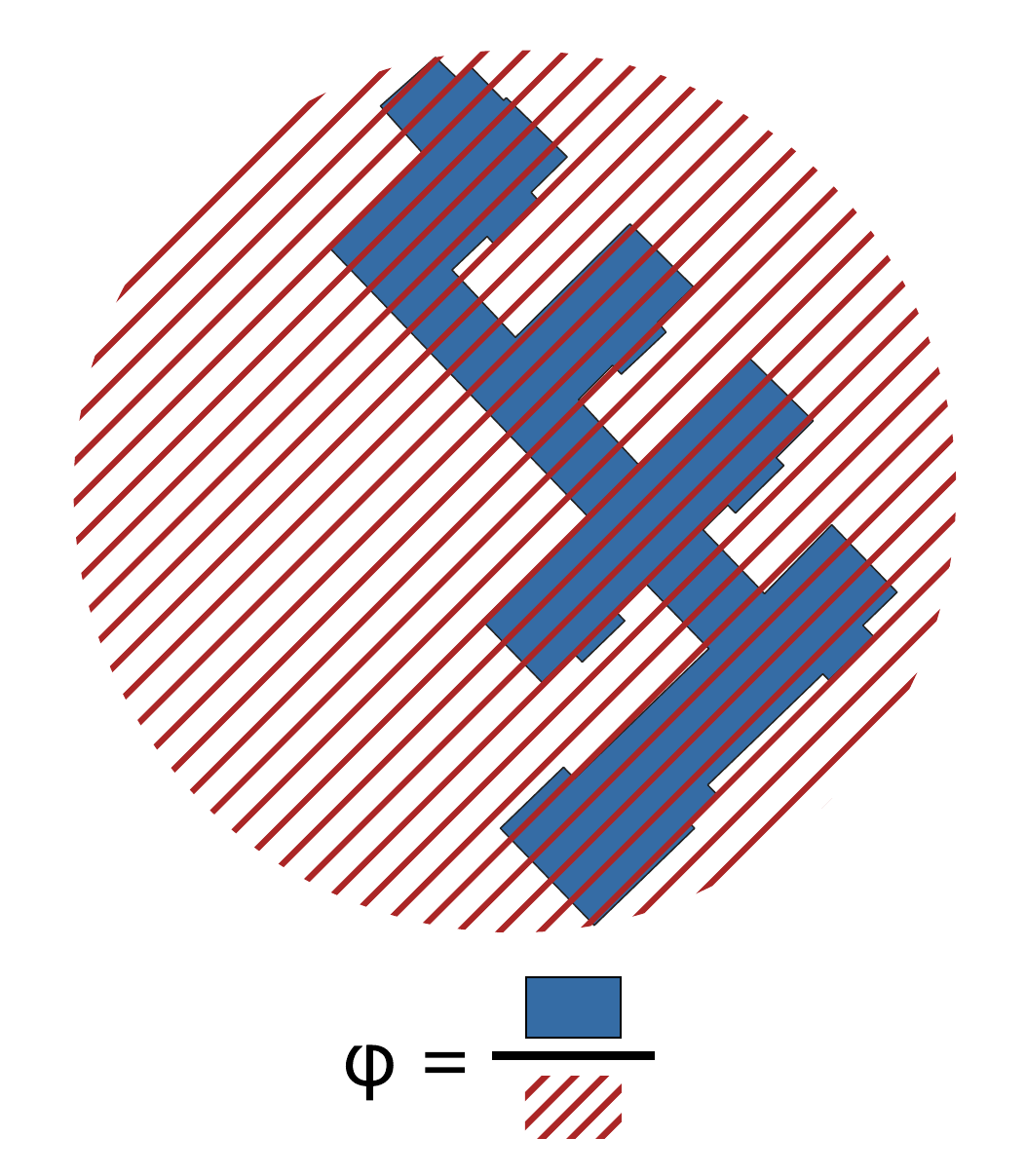} & \includegraphics[scale=0.25]{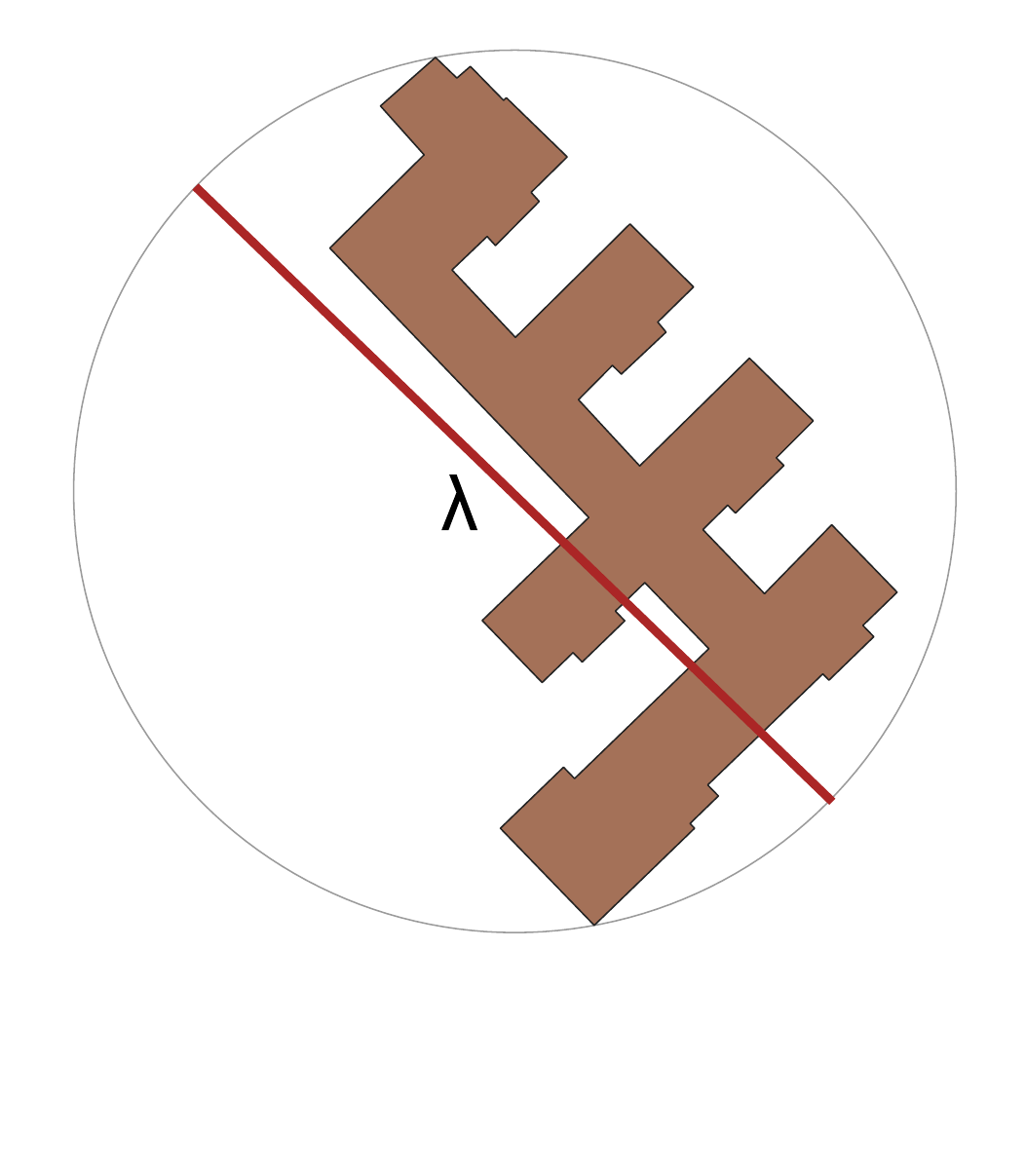} \\
    \specialrule{.1em}{.05em}{.05em}
    \emph{Feature} & \textbf{Elongation} & \textbf{Convexity} \\
    \hline
    \emph{Formula} &
    \begin{equation*}
    \eta =
    \begin{cases}
        \frac{a}{b}, & \text{if } a \leq b \\
        \frac{b}{a}, & \text{if } a > b\\
    \end{cases}
    \end{equation*} \newline
    $a, b:$ sides of minimum bounding box around footprint &
    \begin{equation*}
        \gamma = \frac{\text{Area}(F)}{\text{Area}(H)}
        \label{eq:convexity}
    \end{equation*}  \newline
    $F :$ footprint geometry \newline 
    $H:$ convex hull around footprint\\
    \hline
    \emph{Scale} & [0, 1] & [0, 1]  \\
    \hline
    \emph{Visualization} & \includegraphics[scale=0.25]{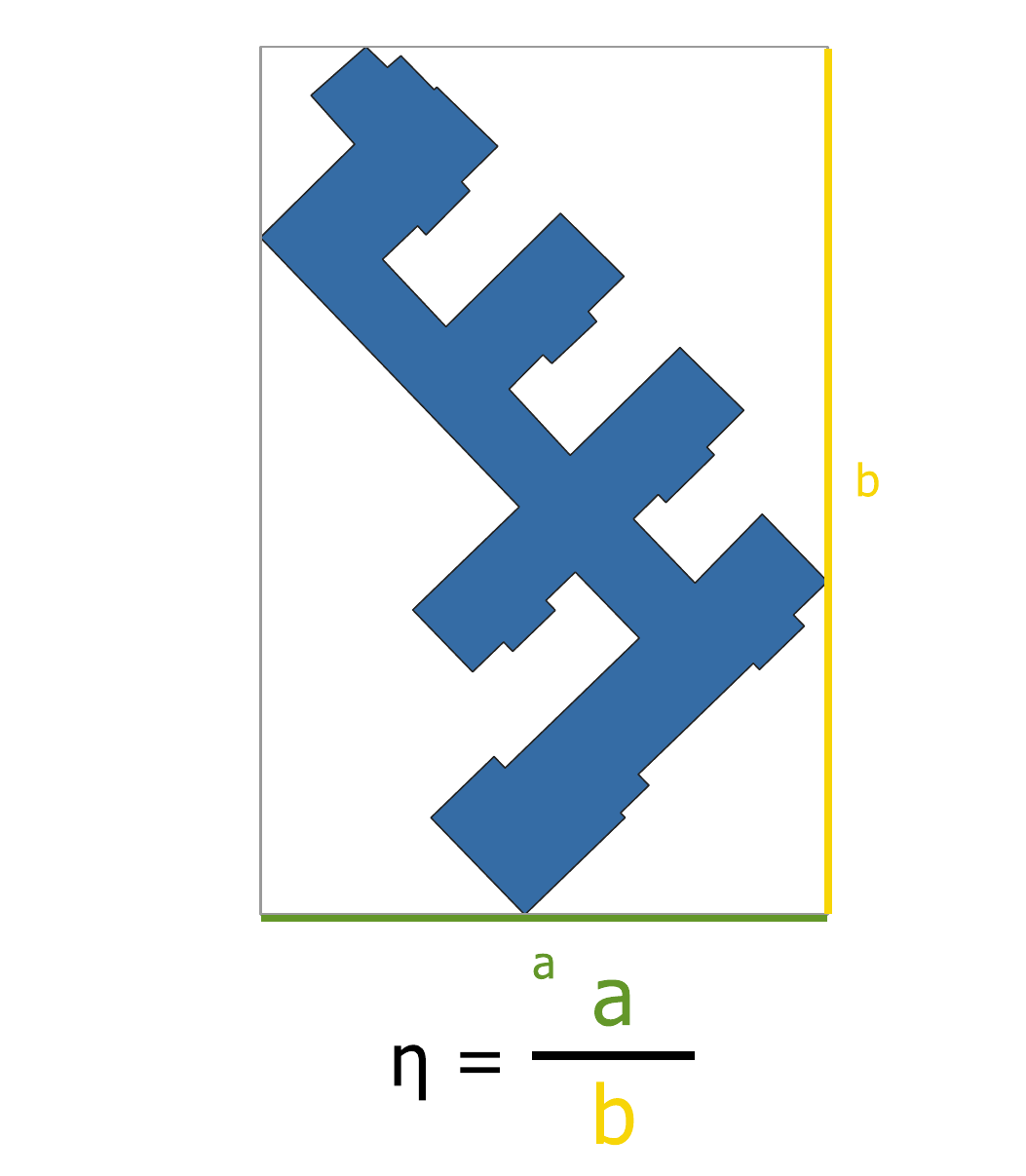} & \includegraphics[scale=0.25]{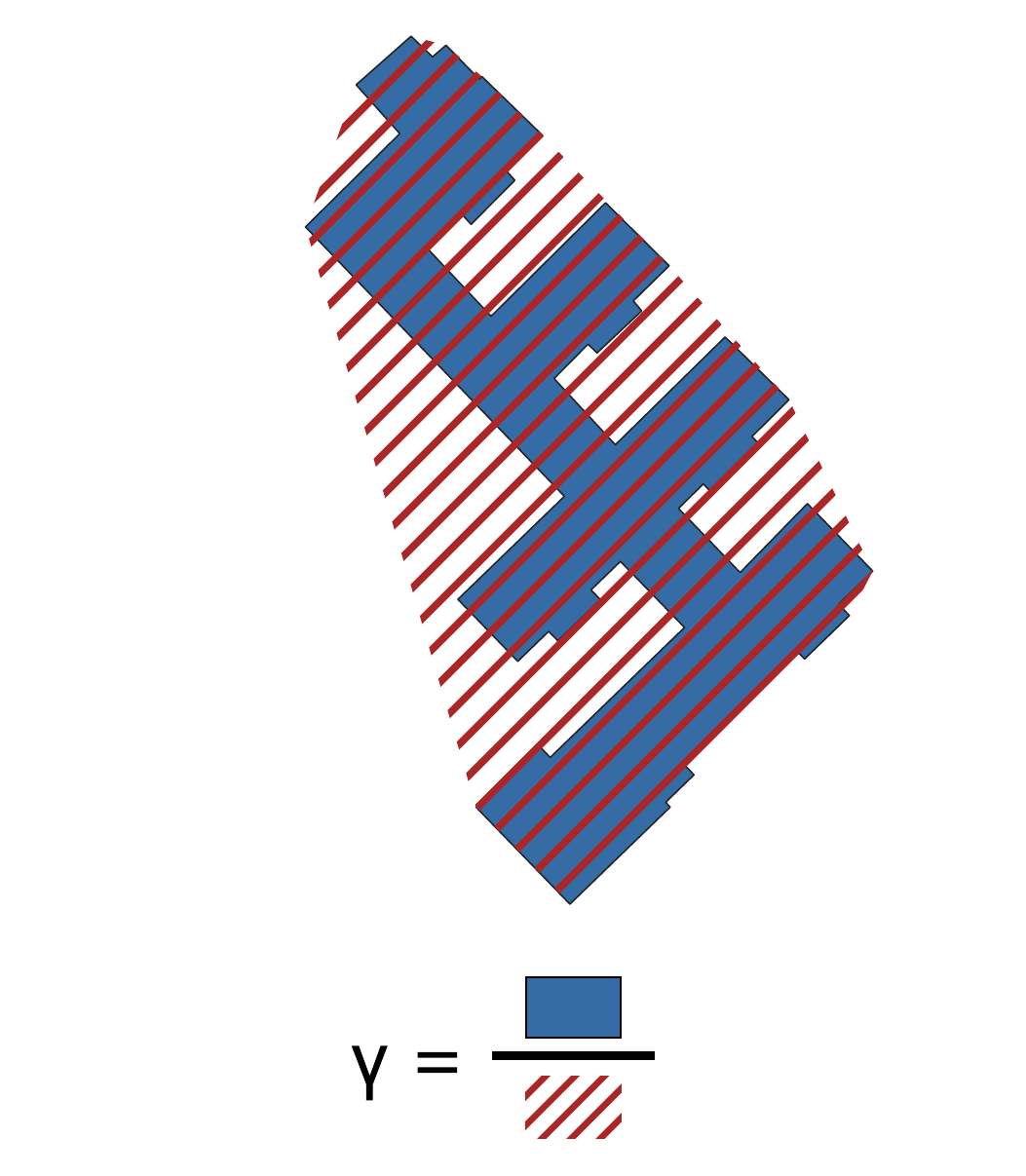} \\
    \specialrule{.1em}{.05em}{.05em}
\end{tabular}}
\label{tab:morphological_features}
\end{table}

\subsubsection{Orientation}

Another shape indicator describes the orientation $\omega$ of the minimum bounding box around the building footprint. $\omega$ is a value between 0° and 45°, a measure for the alignment with the cardinal axes. Buildings whose bounding box is aligned with the cardinal axes have a minimum value of $\omega$, and buildings that are inclined 45° to the side have a maximum value. The orientation is given by

\begin{equation}
\omega = 
    \begin{cases}
            |((azimiuth(p_1, p_2) + 45) \text{ mod } 90) - 45|, & \text{if } a \geq b \\
            |((azimuth(p_1, p_4) + 45) \text{ mod } 90) - 45|, & \text{if } a < b,\\
    \end{cases}
    \label{eq:orientation}
\end{equation}
where $p_1, p_2, p_3, p_4$ are the angle points of the bounding box, $a$ is the rectangle side that connects $p_1$ and $p_2$, and $b$ is the rectangle side that connects $p_1$ and $p_3$. $azimuth(x, y)$ is a function that computes the azimuth angle of $y$ from $x$. We assume that the orientation adds some additional expressiveness to the feature set, especially when features of neighboring buildings are also taken into account.~\citet{Xu20188676} show that orientation plays an important role in such a context as residential buildings are usually regular with each other.

\subsubsection{Number of directly adjacent buildings, shared wall length}

Furthermore, we encode information about whether a building is free-standing or has directly adjacent neighboring buildings. To compute such information, we check if the exterior boundary of a building's footprint intersects with other footprints. We count the number of adjacent buildings and sum the length of the line segments intersecting other buildings. This gives two shape indicators: \emph{number of directly adjacent buildings} and \emph{shared wall length}.

\newpage

\section{Dataset generation}\label{sec:dataset_generation}

\begin{algorithm}
    \caption{Generate localized subgraphs dataset}\label{alg:generate_localized_subgraphs_dataset}
    \begin{algorithmic}
        \Require \\ $subsample\_labeled\_nodes(n_{\text{graphs}})$: function to subsample a random fraction of \newline labeled nodes in the study area to be used as centers for subgraphs\\$n_\text{sub}$: minimum number of buildings a subgraph should contain\\$createbuf(i, r)$: function creating a buffer with metric radius $r$ around node $i$\\ $count(buf)$: function that counts all buildings in the buffer $buf$\\$intersects(a, b)$: function that determines if two polygons $a$ and $b$ intersect\\$building(i)$: building polygon belonging to node $i$.
        \State $\mathcal{G} \gets \{\}$
        \Comment{Start with empty graph dataset}
        \State $\mathcal{S} \gets subsample\_labeled\_nodes(n_\text{graphs})$
        \For{$i \in \mathcal{S}$}
            \State $r \gets 10$
            \Comment{Start with radius of 10 m}
            \State $buf = createbuf(i, r)$
            \While{true}
                \State $n \gets count(buf)$
                \If{$n \geq n_\text{sub}$}
                    \Comment{Buffer contains enough buildings}
                    \State {\textbf{break}}
                \EndIf
                \State $s \gets s + 10$ \Comment{Increase radius in steps of 10 m}
                \State $buf \gets createbuf(i, r)$
            \EndWhile
            \State $\mathcal{V}_i \gets \{j \in \mathcal{V} \ | \ intersects(building(j), buf)\}$
            \Comment{Collect nodes}
            \State $\mathcal{E}_i \gets \{\{j, k\} \ | \ j, k \in \mathcal{V}_i \ \text{ and } \ \exists x\in \mathbb{R}^ 2: x\in \mathcal{R}^{(j)}_i \wedge x\in \mathcal{R}^{(k)}_i \}$
            \Comment{Collect edges}
            \State $\mathcal{G}_i \gets (\mathcal{V}_i, \mathcal{E}_i)$
            \State $\mathcal{G}.add(\mathcal{G}_i)$
            \Comment{Add subgraph to dataset}
        \EndFor \\
        \Return{$\mathcal{G}$}
    \end{algorithmic}
\end{algorithm}

\newpage

\section{Hyperparameters}

\begin{table}[h]
\tbl{Hyperparameters.}
{\begin{tabular}{p{.2\textwidth} p{.08\textwidth} p{.08\textwidth} p{.08\textwidth} p{.08\textwidth} p{.08\textwidth} p{.08\textwidth} p{.08\textwidth}}
    \specialrule{.1em}{.05em}{.05em}
    &\textbf{Decision tree}&\textbf{Random forest}&\textbf{Fully conn. neural net.}&\textbf{Graph conv. net.}&\textbf{Graph \newline SAGE}&\textbf{Graph attention net.}&\textbf{Graph trans.} \\
    \hline
    \textbf{Criterion/loss}&Gini idx.&Gini idx.&Cross-entropy&Cross-entropy&Cross-entropy&Cross-entropy&Cross-entropy\\
    \hline
    \textbf{Optimizer}&-&-&Adam&Adam&Adam&Adam&Adam\\
    \hline
    \textbf{Activation} \newline \textbf{function}&-&-&ReLU&ReLU&ReLU&ReLU&ReLU\\
    \hline
    \textbf{Max. depth}&19&30&-&-&-&-&-\\
    \hline
    \textbf{Num. trees}&-&30&-&-&-&-&-\\
    \hline
    \textbf{Max. features}&-&9&-&-&-&-&-\\
    \hline
    \textbf{Learning rate}&-&-&1e-4&5e-4&1e-4&1e-4&5e-4\\
    \hline
    \textbf{Beta 1 (Adam)}&-&-&0.9&0.9&0.9&0.9&0.9\\
    \hline
    \textbf{Beta 2 (Adam)}&-&-&0.999&0.999&0.999&0.999&0.999\\
    \hline
    \textbf{Batch size}&-&-&1,024&256&256&256&256\\
    \hline
    \textbf{Dropout rate}&-&-&0.35&0.25&0.25&0.25&0.35\\
    \hline
    \textbf{Dropout rate} \newline \textbf{(attention)}&-&-&-&-&-&0&0\\
    \hline
    \textbf{Weight decay}&-&-&1e-5&0&0&0&0\\
    \hline
    \textbf{Num. standard} \newline \textbf{layers}&-&-&3&2&2&2&2\\
    \hline
    \textbf{Num. GNN} \newline \textbf{layers}&-&-&-&2&3&3&4\\
    \hline
    \textbf{Layer sizes}&-&-&1,024&512&1024&512&512\\
    \hline
    \textbf{Attention heads}&-&-&-&-&-&8&8\\
    \hline
    $\zeta$ \textbf{(LeakyReLU} \newline \textbf{negative slope)}&-&-&-&-&-&0.2&-\\
    \hline
    \textbf{Aggregation \newline function}&-&-&-&-&max&-&-\\
    \hline
    \textbf{Residual} \newline \textbf{connections}&-&-&-&-&-&-&no\\
    \hline
    \textbf{$\xi$ (threshold for \newline edge feature \newline scaling)}&-&-&-&500&-&50&50\\
    \hline
    \textbf{Max. epochs}&-&-&200&200&200&200&200\\
    \hline
    $n_\text{sub}$&\multicolumn{7}{c}{20} \\
    \specialrule{.1em}{.05em}{.05em}
\end{tabular}}
\label{tab:hyperparameters}
\end{table}

For the graph attention network and graph transformer, we notice that using multiple attention heads leads to higher performance than using just one -- we use 8 attention heads for both the graph attention network and the graph transformer. Intuitively, each attention head can 'learn' a different concept, in our case, a different type of relationship between buildings. Against this background, a performance gain when using multiple attention heads matches the expectations.

\newpage

\section{Dataset details}

\begin{figure}[h]
\centering
\resizebox*{9cm}{!}{\includegraphics{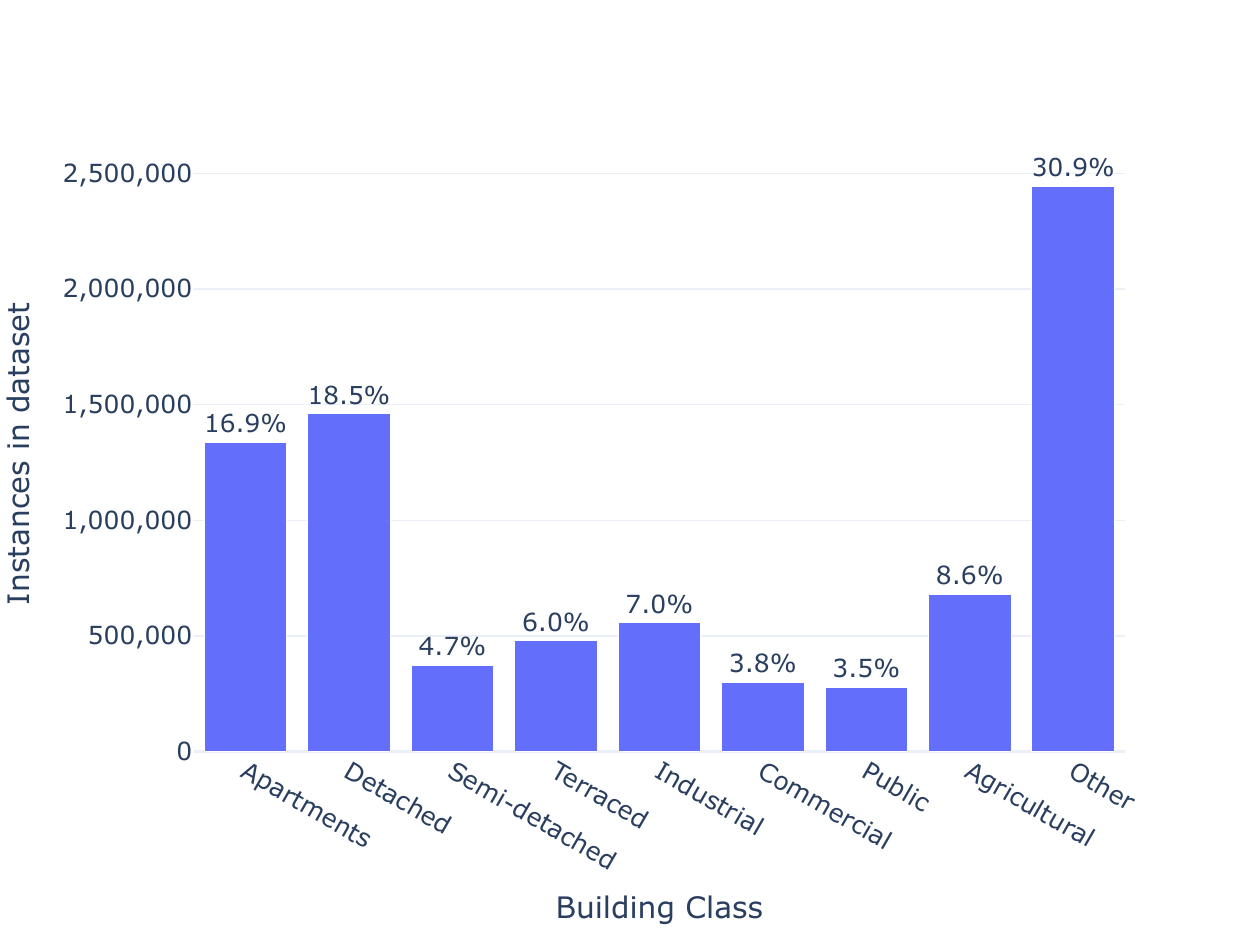}}
\caption{Distribution of building classes.}
\label{fig:class_distribution_histogram}
\end{figure}

\begin{table}[h]
\tbl{Classification scheme of target variable.}
{\begin{tabular}{ p{.25\textwidth}  p{.65\textwidth} }
    \specialrule{.1em}{.05em}{.05em}
    \textbf{Custom building} \newline \textbf{type class} & \textbf{Description}  \\
    \hline
    Apartments & Apartment buildings provide residential accommodation arranged into individual living units, often on several floors. Some apartment buildings have retail outlets on the ground floor. \\
    \hline
    Detached house & A detached house is a free-standing residential building housing a single family. It does not share walls with other buildings. \\
    \hline
    Semi-detached house & A semi-detached house is attached to another house on one side, but the complex does not consist of more than two units. \\
    \hline
    Terraced house & A terraced house is a house in a linear row of more than two residential buildings. \\
    \hline
    Industrial building & Industrial buildings are buildings where goods are produced. Warehouses for the storage of goods are also included in the industrial sector. \\
    \hline
    Commercial building & Commercial buildings mainly deal with services and trade (tertiary sector). The class includes retail stores, supermarkets, offices, hotels, etc. \\
    \hline
    Public building & Public buildings are for the general public and include religious buildings, libraries, townhalls, universities, fire stations, government buildings and government offices, hospitals, kindergartens, educational facilities, museums, police stations, transport facilities, sports centers, etc. \\
    \hline
    Agricultural building & Agricultural buildings include all buildings for agricultural use including barns, cowsheds, and greenhouses. Farm houses do not belong to this category, but are considered detached residential buildings. \\
    \hline
    Other building & Other buildings are mostly outbuildings such as small unmanned service buildings, buildings for storage, and garages. They are not considered very relevant to the energy context. \\
    \specialrule{.1em}{.05em}{.05em}
\end{tabular}}
\label{tab:building_type_classification_scheme}
\end{table}

\begin{table}[h]
\centering
\tbl{Custom classification scheme of building classes.}
{\begin{tabular}{ p{.25\textwidth}  p{.65\textwidth} }
    \specialrule{.1em}{.05em}{.05em}
    \textbf{Custom building} \newline \textbf{class} & \textbf{OSM tags}  \\
    \hline
    Apartments & \emph{apartments, barracks, dormitory}\\
    \hline
    Detached house & \emph{detached, farm} \\
    \hline
    Semi-detached house & \emph{semidetached\_house} \\
    \hline
    Terraced house & \emph{terrace, house} (with additional tag \emph{house=terraced} or \emph{house=terrace})\\
    \hline
    Industrial building & \emph{industrial, electricity, warehouse} \\
    \hline
    Commercial building & \emph{hotel, commercial, kiosk, office, retail, supermarket} \\
    \hline
    Public building & \emph{religious, cathedral, chapel, church, kingdom\_hall, monastery, mosque, presbytery, synagogue, temple, civic, college, fire\_station, government, gatehouse, hospital, kindergarten, museum, public, school, toilets, train\_station, transportation, university, grandstand, riding\_hall, sports\_hall, sports\_centre, stadium} \\
    \hline
    Agricultural building & \emph{barn, cowshed, farm\_auxiliary, greenhouse, slurry\_tank, stable, sty, livestock} \\
    \hline
    Other building & \emph{service, allotment\_house, boathouse, hangar, shed, carport, garage, garages, outbuilding, beach\_hut, container}  \\
    \hline
    Not assigned & \emph{residential, bungalow, cabin, ger, houseboat, static\_caravan, stilt\_house, hut, tree\_house, trullo, shrine, bakehouse, bridge, conservatory, pavilion, parking, transformer\_tower, tech\_cab, water\_tower, storage\_tank, silo, bunker, castle, digester, construction, guardhouse, military, pagoda, quonset\_hut, roof, ruins, tent, tower, windmill} \\
    \specialrule{.1em}{.05em}{.05em}
\end{tabular}}
\label{tab:building_type_mapping_osm}
\end{table}

\begin{figure}[h!]
\centering
\resizebox*{15cm}{!}{\includegraphics{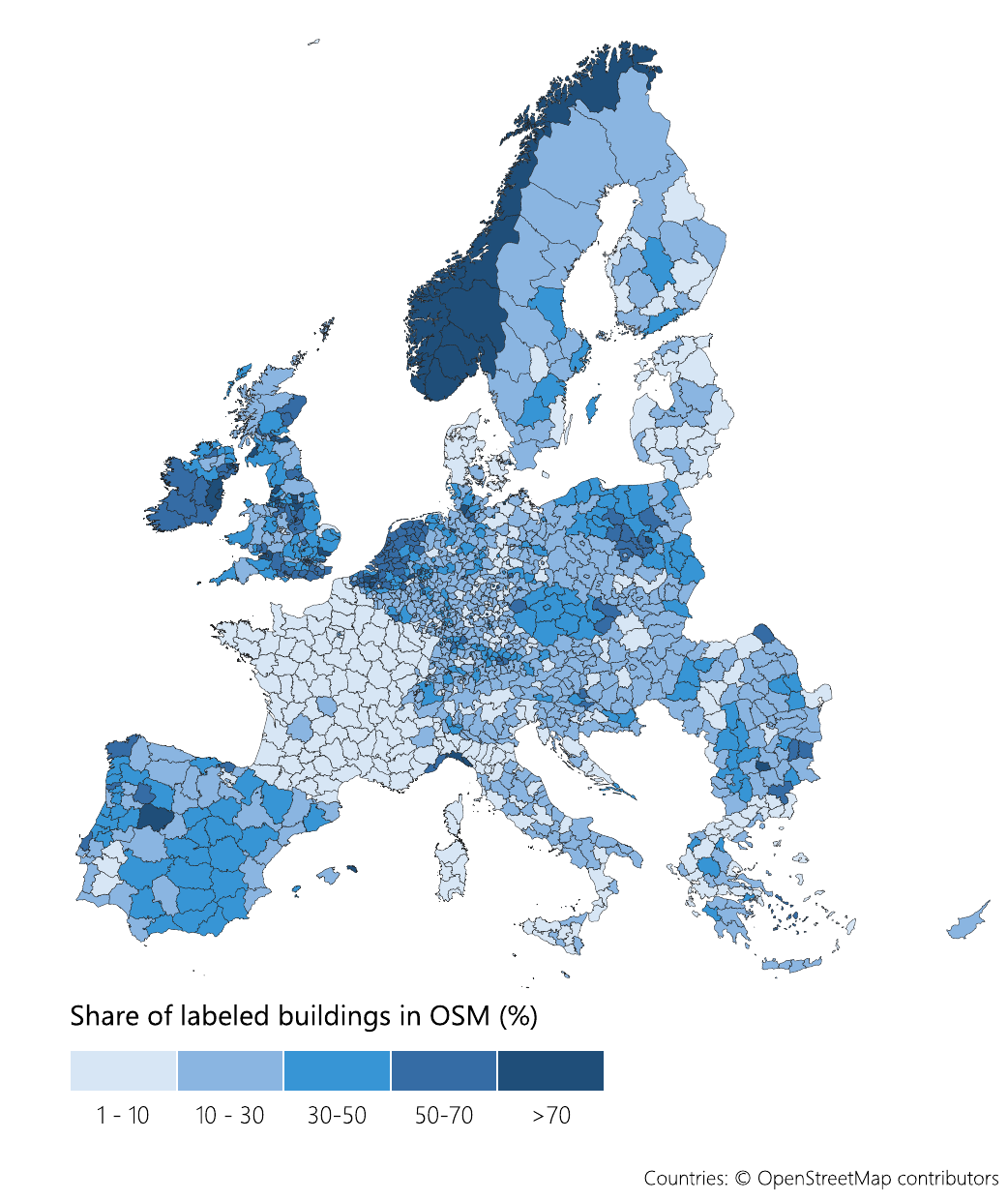}}
\caption{Share of labeled buildings in OSM in different countries of the study area.}
\label{fig:percentage_osm_labeling}
\end{figure}

\newpage

\pagebreak

\begin{table}[h]
\tbl{Custom classification scheme of land use classes.}
{\begin{tabular}
{ p{.45\textwidth}  p{.45\textwidth} }
    \specialrule{.1em}{.05em}{.05em}
    \textbf{Custom land use class} & \textbf{Description}  \\
    \hline
    Continuous urban fabric & \textgreater 80\% of the land surface is covered by buildings, roads
        and artificially surfaced areas \\
    \hline
    Dense or medium density urban fabric & 30-80\% of the land
        surface is covered by buildings, roads, and artificially surfaced areas \\
    \hline
    Low-density urban fabric & 10-30\% of the land surface is covered
        by buildings, roads, and artificially surfaced areas \\
    \hline
    Very low-density urban fabric & \textless 10\% of the land surface is
        covered by buildings, roads, and artificially surfaced areas \\
    \hline
    Isolated structures & structures of buildings/artificially surfaced areas that are not
        part of a larger settlement \\
    \hline
    Industrial/commercial/public/private & areas mainly occupied by manufacturing, trade, financial activities, services, and public activities, as well as private land use \\
    \hline
    Transport & areas mainly occupied by transport infrastructures for road traffic,
        rail networks, airports, river and sea port installations, including their associated
        lands and access infrastructures \\
    \hline
    Mine dump and construction & artificial areas mainly occupied by extractive
        activities, construction sites, and waste dump sites and their associated lands \\
    \hline
    Artificial vegetated areas & areas created for recreational use, for
        instance green or leisure urban parks, sport, and leisure facilities \\
    \hline
    Agricultural & various agricultural areas including arable land, permanent
        crops, and pastures \\
    \hline
    Forests & areas occupied by forests and woodlands \\
    \hline
    Shrub/herbaceous vegetation & includes various areas including
        shrubby areas with heaths, bush and tall herb communities, transitional forest
        development stages, shrubby formation with sparse trees, grasslands, etc. \\
    \hline
    Open spaces & natural areas covered with little or no
        vegetation\\
    \hline
    Wetlands & areas flooded or liable to flooding during the great part of the year
        with specific vegetation coverage made of low shrub, semi-ligneous, or herbaceous
        species\\
        \hline
    Water & inland and marine waters \\
    \specialrule{.1em}{.05em}{.05em}
\end{tabular}}
\label{tab:lc_classification_scheme}
\end{table}

\begin{table}[h]
\centering
\tbl{Mapping from original urban atlas/CORINE land cover classes to custom classes.}
{\begin{tabular}{ p{.4\textwidth}  p{.2\textwidth}  p{.35\textwidth} }
    \specialrule{.1em}{.05em}{.05em}
    \textbf{Custom land use class} & \textbf{CORINE land} \newline \textbf{cover class codes}  & \textbf{Urban atlas} \newline \textbf{class codes}\\
    \hline
    Continuous urban fabric & 111 & 11100 \\
    \hline
    Dense or medium density urban fabric & 112 & 11210, 11220\\
    \hline
    Low-density urban fabric & - & 11230  \\
    \hline
    Very low-density urban fabric & - & 11240  \\
    \hline
    Isolated structures & - & 11300  \\
    \hline
    Industrial/commercial/public/private & 121 & 12100 \\
    \hline
    Transport & 122--124 & 12210, 12220, 12230, 12300, 12400 \\
    \hline
    Mine dump and \newline construction & 131--133 & 13100, 13300, 13400 \\
    \hline
    Artificial vegetated areas & 141, 142 & 14100, 14200\\
    \hline
    Agricultural & 211--213, \newline 221--223, \newline 231, \newline 241--244 & 21000, 22000, 23000, 24000, 25000 \\
    \hline
    Forests & 311--313 & 31000 \\
    \hline
    Shrub/herbaceous \newline vegetation & 321--324 & 32000 \\
    \hline
    Open spaces & 331--335 & 33000 \\
    \hline
    Wetlands & 411, 412, \newline 421--423 & 40000 \\
        \hline
    Water & 511, 512, \newline 521--523 & 50000 \\
    \specialrule{.1em}{.05em}{.05em}
\end{tabular}}
\label{tab:lc_mapping}
\end{table}

\end{document}